\documentclass{article}

\usepackage{microtype}
\usepackage{graphicx}
\usepackage{booktabs} 
\usepackage{subfig}
\usepackage{multirow}
\usepackage{amsmath}
\usepackage{amsfonts}
\usepackage{amssymb}

\usepackage{bm} 
\renewcommand{\mathbf}[1]{\bm{#1}} 

\newcommand{\R}{\mathbb{R}}

\usepackage{hyperref}

\usepackage[accepted]{icml2021}

\icmltitlerunning{Improving Molecular Modeling with Geometric GNNs: an Empirical Study}

\begin{document}

\twocolumn[
\icmltitle{Improving Molecular Modeling with Geometric GNNs: an Empirical Study}



\icmlsetsymbol{equal}{*}

\begin{icmlauthorlist}
\icmlauthor{Ali Ramlaoui}{equal,cs,ens}
\icmlauthor{Théo Saulus}{equal,cs,ens}
\icmlauthor{Basile Terver}{equal,x,ens}
\\
\icmlauthor{Victor Schmidt}{udem,mila,entalpic}
\icmlauthor{David Rolnick}{mila,mcgill}
\icmlauthor{Fragkiskos D. Malliaros}{cs,inria}
\icmlauthor{Alexandre Duval}{cs,inria,mila,entalpic}
\end{icmlauthorlist}

\icmlaffiliation{cs}{Université Paris-Saclay, CentraleSupélec}
\icmlaffiliation{ens}{École Normale Supérieure Paris-Saclay}
\icmlaffiliation{x}{École Polytechnique, IP Paris}
\icmlaffiliation{mila}{Mila - Quebec AI Institute}
\icmlaffiliation{mcgill}{McGill University}
\icmlaffiliation{entalpic}{Entalpic}
\icmlaffiliation{inria}{Inria Saclay}
\icmlaffiliation{udem}{Université de Montréal}

\icmlcorrespondingauthor{Ali Ramlaoui}{ali.ramlaoui@student-cs.fr}
\icmlcorrespondingauthor{Théo Saulus}{theo.saulus@student-cs.fr}
\icmlcorrespondingauthor{Basile Terver}{basile.terver@polytechnique.edu}

\icmlkeywords{Machine Learning, ICML}

\vskip 0.3in
]



\printAffiliationsAndNotice{\icmlEqualContribution} 

\begin{abstract}
Rapid advancements in machine learning (ML) are transforming materials science by significantly speeding up material property calculations. However, the proliferation of ML approaches has made it challenging for scientists to keep up with the most promising techniques. This paper presents an empirical study on Geometric Graph Neural Networks for 3D atomic systems, focusing on the impact on performance, scalability, and symmetry enforcement of different (1) canonicalization methods, (2) graph creation strategies, and (3) auxiliary tasks. Our findings and insights aim to guide researchers in selecting optimal modeling components for molecular modeling tasks. Our code is available at \url{https://github.com/RolnickLab/ocp}.
\end{abstract}

\section{Introduction}
\label{Introduction}

The field of computational materials science has witnessed an increasing interest in recent years, with the explosion of machine learning approaches to model material properties at the quantum scale. This is possible thanks to the release of large-scale datasets such as OC20 \cite{chanussot2020open} and QM7-X \cite{hoja2021qm7x}, which contain millions of molecular structures along with various properties (forces, energy, band gap) computed using quantum mechanical simulations involving Density Functional Theory (DFT) \cite{kohn1996density}. 

ML models have been trained to approximate DFT, thus constituting an even faster proxy to the Schrödinger equation, modeling atomic interactions and systems' behavior. They enable the calculation of material properties in seconds instead of hours or days, offering the potential to accelerate scientific discovery via high-throughput screening of novel materials \cite{batatia2023foundation, merchant2023scaling}. However, the explosion of ML approaches in recent years makes it hard to keep up with promising techniques for scientists in the field. While some surveys have attempted to structure the different categories of ML approaches \cite{han2022geometrically, duval2024hitchhiker}, these do not focus on empirical evaluation.

In this paper, we propose an empirical study of some key modeling aspects of Geometric GNNs for 3D atomic systems. Specifically, we investigate the impact of 1) the recent canonicalization methods used to enforce or approximate Euclidean symmetries, 2) the graph creation step when modeling a 3D atomic system, and 3) adding several auxiliary tasks. We focus on the OC20 dataset modeling the relaxed adsorption energy of an adsorbate-catalyst system. 
We hope that the conclusions and insights drawn from our experiments will benefit the community, making it possible to quickly choose the right modeling component. 

\section{Choice of Canonicalization} \label{sec:canonicalization}
A function $f : \mathcal{X} \rightarrow \mathcal{X}$ is said to be equivariant with respect to a transformation $t$ if $\forall x \in \mathcal{X}, f(t(x))=t(f(x))$. In particular, $f$ is $E(3)$-equivariant if it is equivariant to rotations, translations, and reflections.
$E(3)$-equivariance is a desirable property in molecular modeling to learn representations that are best suited for physically meaningful tasks, such as force predictions on atoms (e.g., S2EF task of OC20). This can be enforced in the architecture of the model during the message passing steps by using equivariant features of the input’s representation \cite{painn, batatia2023mace, liao2023equiformer}, which comes at the cost of expensive feature computations. A recent alternative to these equivariant architectures lies in unconstrained GNNs, which do not enforce $E(3)$-equivariance by model design but instead with a coordinate-preprocessing step referred to as canonicalization \cite{hu2021forcenet, duval2023faenet, pozdnyakov2024smooth}. This process grants unconstrained GNNs with greater flexibility, scalability, and often expressivity \cite{duval2024hitchhiker}. Commonly, it involves projecting (e.g., with PCA or an equivariant network) the input atomic system into a canonical space such that every Euclidean transformation of the same system gets mapped to the same canonical representation, i.e.~handling symmetries in the data pre-processing.

Since these canonicalization methods are all recent and no comparison has yet been drawn, we benchmark the proposed approaches on QM9 and OC20 tasks, evaluating their impact on performance and symmetry enforcement.

\subsection{Canonicalization for direct predictions}

In this subsection, we evaluate several canonicalization procedures with the FAENet backbone architecture \cite{duval2023faenet}, a powerful unconstrained GNN 
which is not equivariant by itself.
We assume familiarity of the reader with these approaches but provide a description in Appendix \ref{appendix:cano}.
\begin{itemize}
    \item Vector Neurons Network (VNN) \cite{deng2021vector} using the VN re-implementation of PointNet \cite{qi2017pointnet} and DGCNN \cite{wang2019dgcnn}. This class of canonicalization networks is $SO(3)$-equivariant by design and is applied to obtain a canonical representation of the data 
    following the method of \citet{kaba2023equivariance}.
    \item Stochastic Frame Averaging (SFA) \cite{duval2023faenet}, designed to avoid averaging the predictions over $8$ elements of the frame as required by Frame Averaging \cite{puny2021frame}. Instead, we sample one canonical orientation at random at each epoch, similar to data augmentation on a small and complete set.
    \item A novel method denoted SFA+SignNet, where we propose to map the several projection matrices of SFA to a single one using a sign-equivariant network proposed in SignNet \cite{lim2023signinv, lim2023signequiv}. The rationale is to handle the sign ambiguity problem of PCA that exists in SFA with a small dedicated network to get a unique canonical representation of Euclidean transformations. We propose two implementations of SignNet, either using VNNs to have a perfect $E(3)$-equivariance when combined with SFA or using MLPs without theoretical guarantees.
\end{itemize}
 

For VNNs and SFA+SignNet methods, we test both training and freezing the weights of the Canonicalization Network (CN). We report performance on the OC20 IS2RE, OC20 S2EF, and QM9 
in Tables \ref{table:lcf_is2re_summary}, \ref{table:lcf_is2re_acc}, \ref{table:lcf_is2re_perf}, \ref{table:lcf_s2ef_acc}, \ref{table:lcf_s2ef_perf}, and \ref{table:lcf_qm9_acc}. 



In our OC20 IS2RE experiment, we found there are almost no differences between the various canonicalization methods, with MAE of 594, 598, and 593 for SFA, VN-PointNet, and VN-DGCNN. Specially, non-exact canonicalizations (SFA and SFA+SignNet) demonstrate equal or better MAE than perfectly equivariant methods (e.g., VN-based). This suggests that heuristics approximation of equivariance should be sufficient in some practical applications like OC20. This is aligned with what \citet{duval2023faenet} suggested when showing that SFA outperforms exact Frame Averaging in terms of downstream performance. 

In terms of symmetry enforcement, non-exact methods are surprisingly nearly as effective as fully invariant methods, suggesting that the FAENet backbone implicitly learns to handle symmetries.

Regarding exact canonicalization methods, we observe that training or not the network and swapping one method for another has little impact on model performance. 
This tends to indicate that the canonical networks's ability to introduce equivariance is more critical than the choice of the canonicalization method.

\begin{table}[t]
\centering
\resizebox{0.5\textwidth}{!}{
    \begin{tabular}{l|c|cc|c}
    Canonicalization & Cano. trained & avg. MAE & EwT (ID) & 3D Rotation \\ 
    & parameters & (meV) $\downarrow$ & (\%) $\uparrow$ & Invariance $\downarrow$ \\
    \hline
    SFA & 0 & 594 & 4.40 & $1.30 \cdot 10^{-2}$ \\
    (U) SFA+MLP-SignNet & 0 & \textbf{580} & 4.48 & $9.71 \cdot 10^{-2}$\\
    (T) SFA+MLP-SignNet & 454 & 583 & 4.46 & $4.00 \cdot 10^{-2}$ \\
    (U) SFA+VN-SignNet & 0 & 592 & \textbf{4.69} & $7.58 \cdot 10^{-3}$\\
    (T) SFA+VN-SignNet & 2,620 & 599 & 4.25 & $2.57 \cdot 10^{-2}$ \\
    \hline
    (U) VN-Pointnet & 0 & 605 & 4.09 & $4.62 \cdot 10^{-3}$ \\ 
    (T) VN-Pointnet & 1,310 & 598 & 4.12 & $\mathbf{3.80 \cdot 10^{-3}}$ \\ 
    (U) VN-DGCNN & 0 & 600 & 4.31 & $3.11 \cdot 10^{-2}$ \\ 
    (T) VN-DGCNN & 663,804 & 593 & 4.42 & $9.10 \cdot 10^{-3}$ \\ 
    \end{tabular}
}
\caption{Invariance comparison of canonicalization methods on OC20 IS2RE dataset. (U) (resp. (T)) indicates an untrained (resp. trained) canonicalization network. FAENet backbone has 4,147,731 parameters. More details in Tables \ref{table:lcf_is2re_acc} and \ref{table:lcf_is2re_perf}.}
\label{table:lcf_is2re_summary}
\end{table}

\subsection{Canonicalization for relaxed IS2RE}
\label{sec:relaxed_is2re}
Previous work showed that solving the IS2RE task yields better results by performing relaxed energy predictions rather than direct energy estimation \citep{liao2023equiformerv2}. Here, we evaluate whether canonicalization methods also perform well at relaxing a trajectory. 
Table~\ref{table:relaxed_s2ef} reports the performances of FAENet with multiple symmetry-preserving methods, with the invariant SchNet model 
and direct IS2RE acting as baseline. 
Our findings suggest that relaxed IS2RE predictions are competitive with direct IS2RE predictions and are interesting directions to explore for improving molecular property predictions with these architectures. Moreover, a potential explanation as to why the relaxations do not yield significant improvements over direct IS2RE may involve the approximate equivariance or the lack of continuity in some of these canonicalization methods, as pointed out in \citet{dym2024equivariant}, which may hamper accurate and smooth relaxations trajectories. This is mainly true for SFA, where a frame is randomly picked at each step of the relaxation, meaning that the canonical inputs can also be far from each other.
Lastly, note that exact equivariant methods for relaxed IS2RE predictions such as the untrained VN-PointNet (UTPN) implementation from \cite{kaba2023equivariance} yields better MAE than the approximate equivariance module SFA. Still, during our experiments we found that with further focus on the accuracy of S2EF models, canonicalization methods used to enforce symmetries can become more appealing using iterative relaxation methods.
\begin{table}[t]
\centering
\resizebox{0.5\textwidth}{!}{
    \begin{tabular}{l|cc|cc}
    & \multicolumn{2}{c|}{IS2RE} & \multicolumn{2}{c}{IS2RS} \\
    Model & EwT (\%) $\uparrow$& MAE (eV) $\downarrow$& DwT (\%) $\uparrow$& Pos. MAE $\downarrow$\\
    \hline
        FAENet (Direct) & 4.05&  0.551 &  - &  - \\
        FAENet (SFA)& 4.92 &  0.587 &  31.1&  0.390 \\
        FAENet (UTPN) & 5.64 & 0.560 & 33.7 & 0.381 \\
        SchNet & 1.89 & 0.912 & 15.0 & 0.461 \\
    \end{tabular}
    }
    \caption{Results on the IS2RE and IS2RS tasks for the VAL-ID validation dataset of OC20 using iterative relaxations. The S2EF models' results are reported in Appendix~\ref{sec:app_relax}. Note that these results can only be obtained by keeping the tag 0 atoms of the subsurface as they are important for relaxations.
    }
\label{table:relaxed_s2ef}
\end{table}

\section{Graph Creation Study}
For large molecular structures, electrostatic long-range interactions are non-negligible components of the system's dynamics \cite{dimenet}. 
While various methods have tried to model long-range interactions between far-away atoms, they often suffer from the over-smoothing effect when increasing the number of interaction layers
\citep{liao2023equiformer}. As a result, models where the geometric graph is modelled using a cutoff distance between neighbour atoms have been shown to work best. Since properly handling these interactions is essential to accurately simulate the system, we explore in this section the impact of this creation step and the rewiring strategies.

\subsection{Graph cutoff and rewiring}

First, we vary the cutoff distance\footnote{When representing the 3D point cloud with a graph, we create an edge between any two atoms if their are within a fixed cutoff distance, and no edge otherwise.} used during the creation of the graph to check whether linking more atoms with each other helps in learning these couplings. We report the results for the FAENet model on the OC20 IS2RE task in Table \ref{table:cutoff}. 

\noindent
A small cutoff of 1.0 \AA \ leads to the weakest performance, which makes sense since the associated graph is almost empty and nodes are isolated, i.e. messages can not pass correctly. A large cutoff or a complete graph also leads to poor performance despite every atomic interaction being modeled. Within an intermediary range of cutoff values, the model achieves optimal learning with computational efficiency. This is in accordance with past observations where the locality bias of GNN models seemed to fit really well with atomic system property prediction tasks. 
Thus, although the creation of the graph through a well-chosen cutoff is important, the margin for fine-tuning this parameter is large enough. 

The fact that better graphs are adapted to GNN functioning rather than the precise modeling of the situation, where all atoms interact with each other, echoes what \citet{duval2022phast} have stated. Indeed, they showed that to fit the GNN message passing scheme, removing repeating subsurface atoms of the adsorbate did not affect the model performance on IS2RE tasks. Similarly to having a moderate cutoff, the performance improvements of such a strategy are decisive for scalability. As an empirical example, we tried to run an EquiformerV2 \citep{liao2023equiformerv2} model for IS2RE with the remove-tag-0 method on an 80GB A100 GPU, leading to a 5$\times$ speed-up with no performance loss and proving the relevance of this technique even on very large models.

\begin{table}[t]
\centering
\resizebox{0.9\linewidth}{!}{
\begin{tabular}{l|cc}
& \multicolumn{2}{c}{ID} \\
Model & EwT (\%) $\uparrow$& MAE (eV) $\downarrow$\\
\hline
Cutoff 30 - Max neighbours 40 & 2.65 & 0.697 \\
Cutoff 20 - Max. neighbours 40 & 3.08& 0.673 \\
Cutoff 20 - Max. neighbours 10 & 2.25& 0.768 \\
Cutoff 10 - Max. neighbours 50 & 4.17& 0.553 \\
Cutoff 10 - Max. neighbours 10 & 4.49& 0.553 \\
Cutoff 6 - Max. neighbours 40 & 4.31 & 0.553  \\
Cutoff 1 - Max. neighbours 40 & 1.35 & 1.069 
\end{tabular}
}
\caption{Impact of the cutoff on the performance of FAENet on the OC20 IS2RE task. Full Table in Appendix~\ref{sec:app_cutoff}.}
\label{table:cutoff}
\end{table}


\subsection{Ewald-based long range message passing}
\label{sec:ewald}
Since a small value for the cutoff seems to be the most interesting one, we want to model the long interactions differently than adding links between all atoms.
Ewald-based Message Passing (EMP) \cite{kosmala2023ewald} is introduced in this perspective. It incorporates a physics-based prior in the architecture to model the long-range electrostatic potential via a nonlocal Fourier space scheme, drawing edges based on a cutoff on frequency instead of distances.

\begin{figure}[ht]
    \centering
    \subfloat[SchNet without Ewald]{\includegraphics[width=0.495\linewidth]{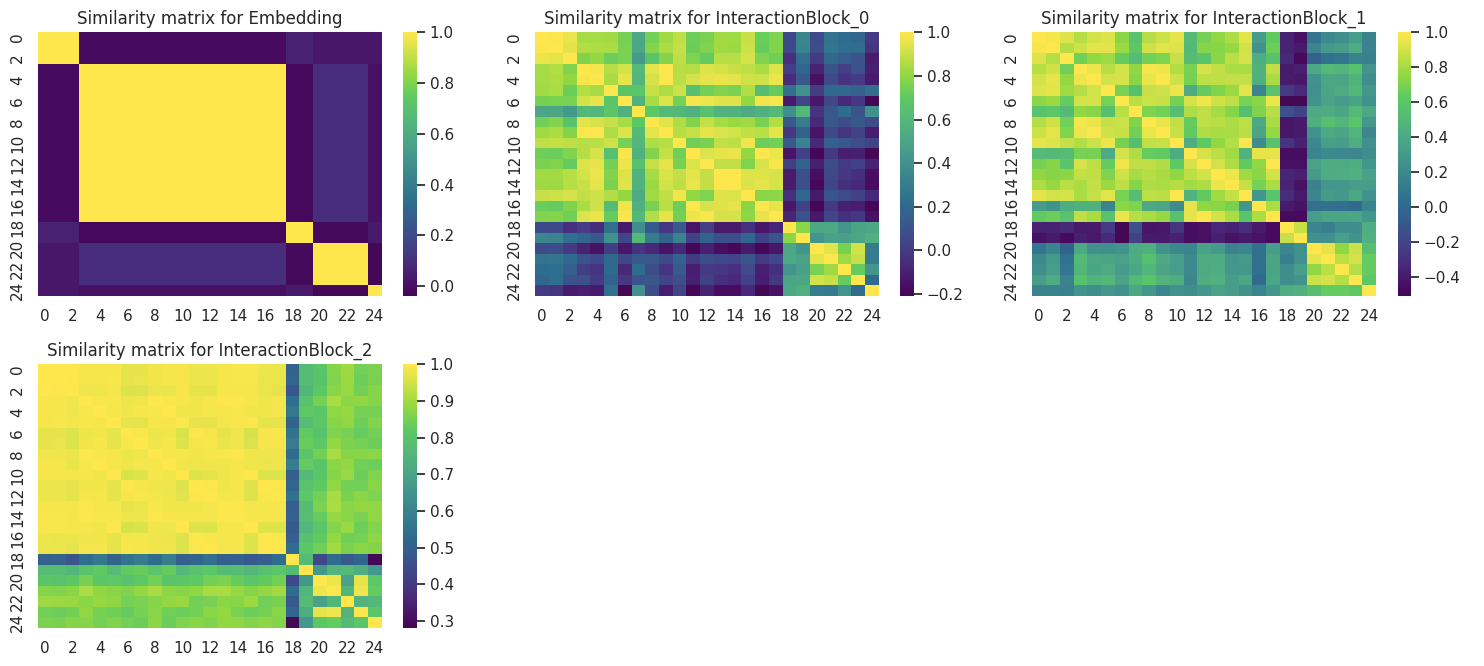}}
    \hfill
    \subfloat[FAENet without Ewald]{\includegraphics[width=0.495\linewidth]{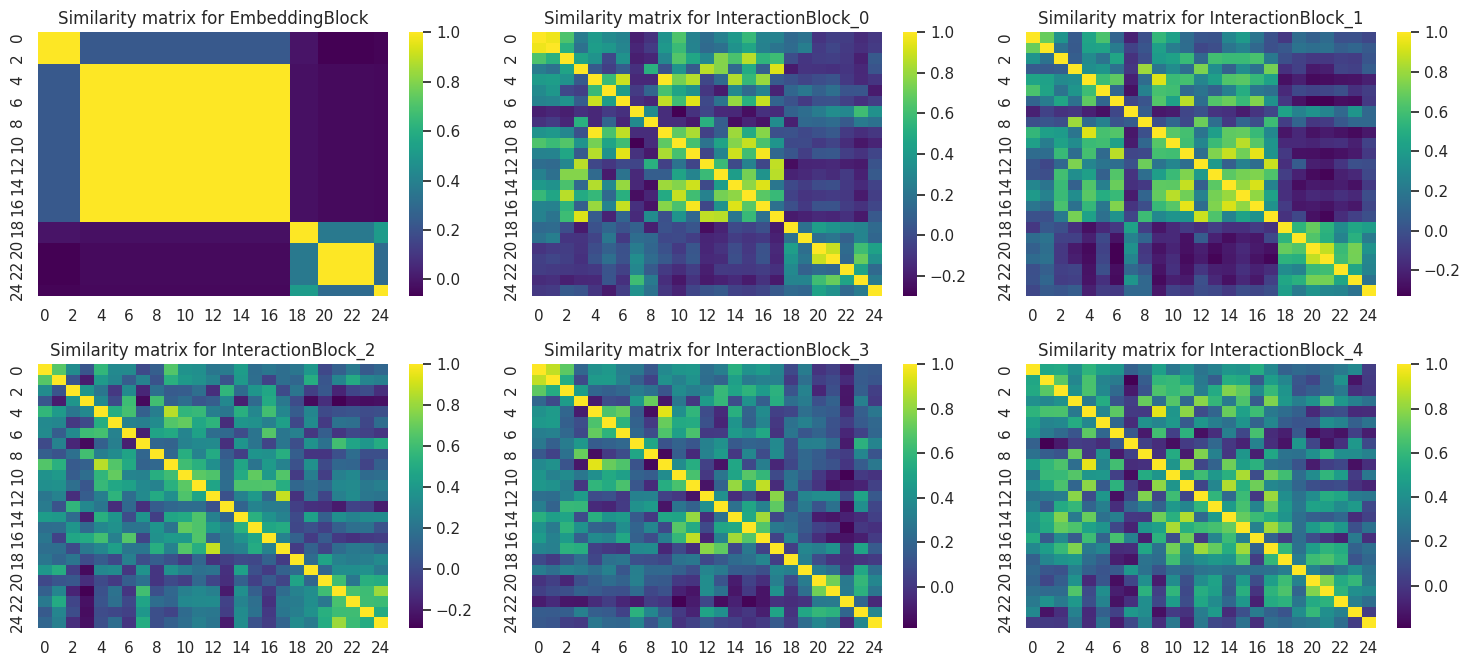}}\\
    \subfloat[SchNet with Ewald]{\includegraphics[width=0.495\linewidth]{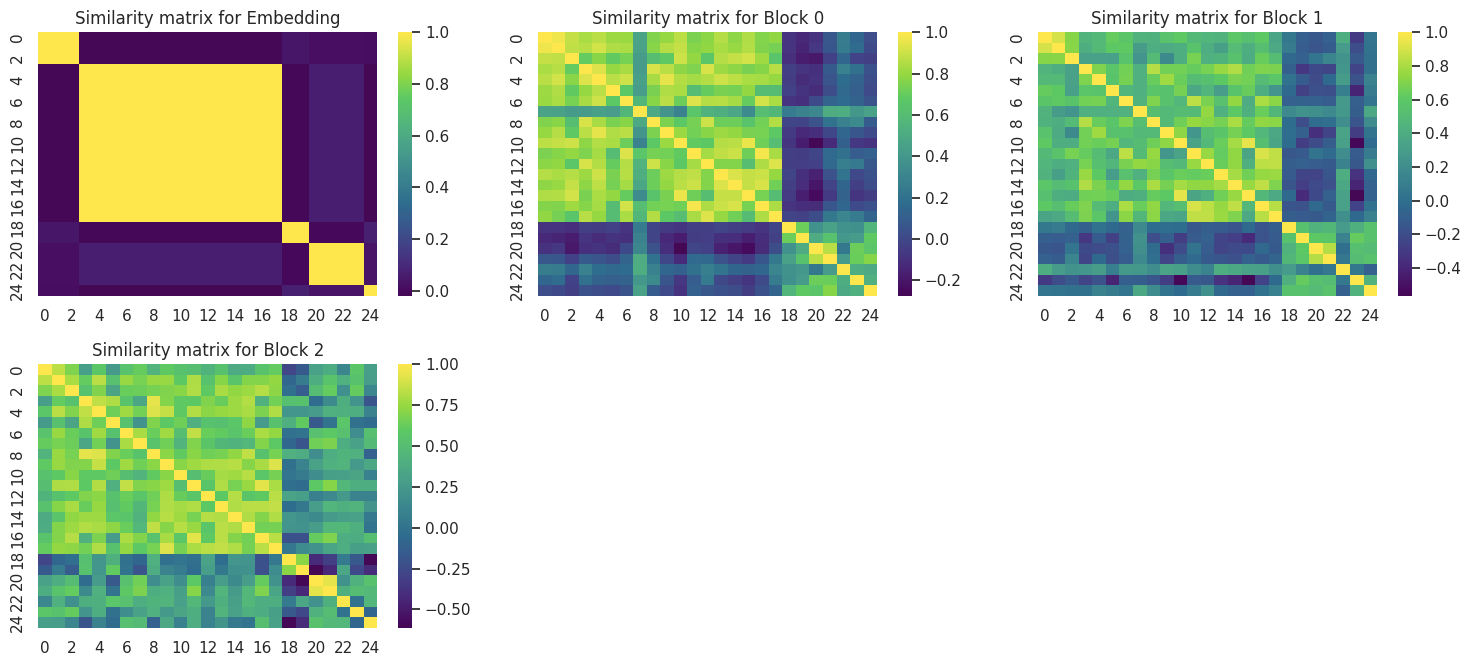}}
    \hfill
    \subfloat[FAENet with Ewald]{\includegraphics[width=0.495\linewidth]{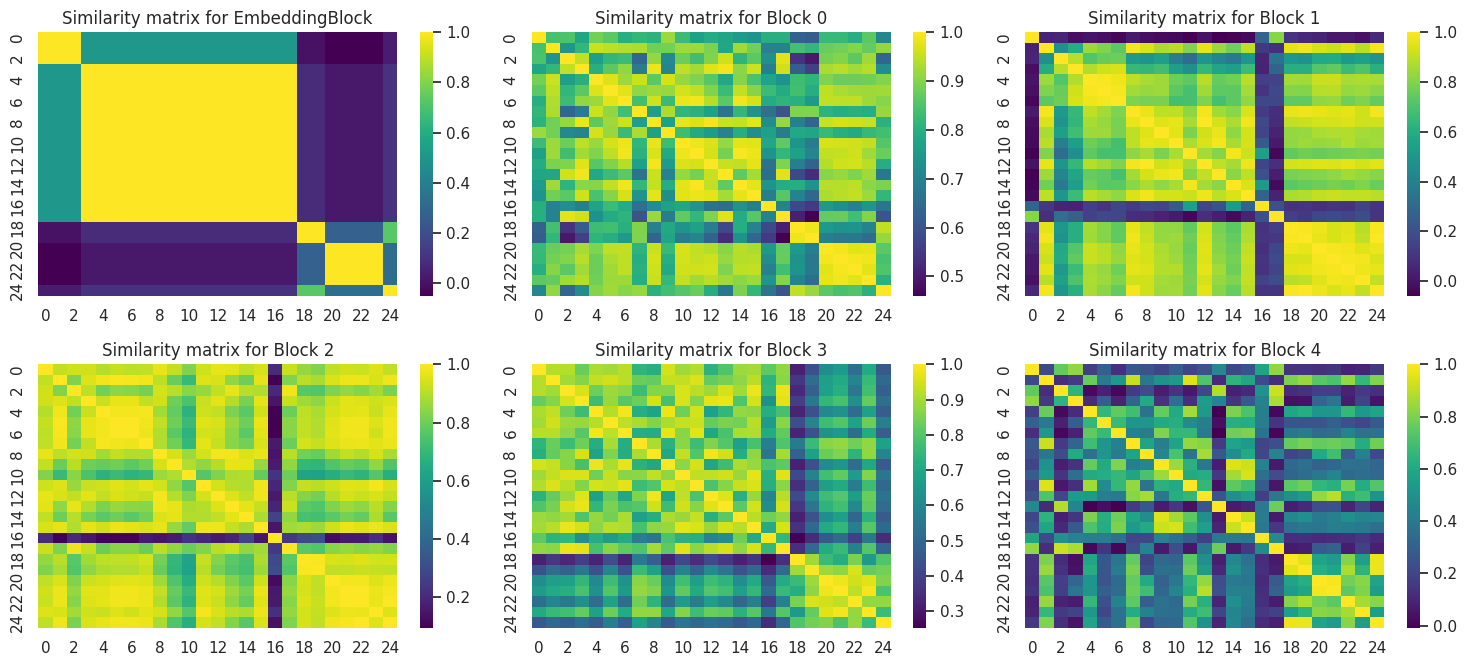}}
    \caption{Similarity matrix of the embeddings of the atoms of a randomly picked system for different interaction blocks from the training set of OC20. The same system is used every time to be able to compare the different results.}
    \label{fig:ewald_embeddings}
\end{figure}

Our experiments, given in Table~\ref{table:ewald}, show that EMP is interesting for an invariant method like SchNet \cite{schutt2017schnet}, which limits its geometric information to atom pairwise distances. However, EMP does not benefit more advanced GNNs like FAENet. To understand why, we plot in Figure~\ref{fig:ewald_embeddings} the cosine similarity between the embeddings throughout interaction layers. We observe that SchNet and FAENet learn very different representations. 
Indeed, while the representation of each atom in SchNet tends to be similar to nearby atoms, FAENet is able to give very different embeddings for them. A potential explanation could be that because FAENet is a more expressive model, the propagated messages are less constrained and thus can lead to very diverse atom representations without Ewald. On the other hand, embeddings of SchNet only become diversified using Ewald, maybe because incorporating longer messages helps create more distinct representations. 
More plots 
can be found in Appendix~\ref{sec:app_ewald}. 
One interesting takeaway is that EMP helps take into account long-range interactions for simple models and GNNs with symmetry-constrained layers \citep{kosmala2023ewald} but is less efficient on expressive models.

\begin{table}[t]
\centering
\resizebox{1\linewidth}{!}{
    \begin{tabular}{l|cc}
    & \multicolumn{2}{c}{ID} \\
    Model & EwT (\%) $\uparrow$& MAE (eV) $\downarrow$\\
    \hline
        FAENet (Graph Rewiring) & 4.05&  0.551 \\
        FAENet (Graph Rewiring) + Ewald & 4.12  & 0.562 \\
        FAENet (No Graph Rewiring) & 4.54 &  0.544 \\
        FAENet (No Graph Rewiring) + Ewald & 4.11 & 0.556 \\
        SchNet (Graph Rewiring) & 3.18 &  0.641 \\
        SchNet (Graph Rewiring) + Ewald & 3.54 & 0.604 \\
        SchNet (No Graph Rewiring) & 2.93&  0.654 \\
        SchNet (No Graph Rewiring) + Ewald & 3.48  & 0.597 
    \end{tabular}
    }
    \caption{Comparison of the performances of FAENet and SchNet with and without Ewald Message Passing on the IS2RE task. 
    Full table and results for the QM9 dataset in Appendix~\ref{sec:app_ewald}.}
\label{table:ewald}
\end{table}

\section{Auxiliary Tasks} \label{sec:noisy_nodes}
In this section, we study how to leverage other tasks to improve the performance of FAENet on IS2RE.

\subsection{Noisy nodes IS2RS auxiliary task}
First, we recall that increasing the number of interaction blocks above 8 does not yield better performance for most geometric GNNs, with a dramatic loss of information after 14 layers in the classical FAENet IS2RE setup, as illustrated in Table~\ref{table:is2re_top_comp_interactions_fixed}.
To address the oversmoothing issue, \citet{godwin2022simple} propose to use noisy regularisation, introducing their method called Noisy Nodes. It consists of adding an auxiliary node-level denoising task that encourages diversity in the latent representations of the nodes (more details in \ref{Noisy Nodes appendix implementation}).

Here, we implement Noisy Nodes for the IS2RE task, which 
is done by adding a position decoding head running in parallel to the original energy prediction head. Our implementation takes inspiration from other state-of-the-art models such as EquiformerV1 \cite{liao2023equiformer}, which benefit from using Noisy Nodes on the IS2RE task (more details in \ref{Noisy Nodes appendix implementation}).

\begin{figure}[t]
    \centering
    \includegraphics[width=\linewidth]{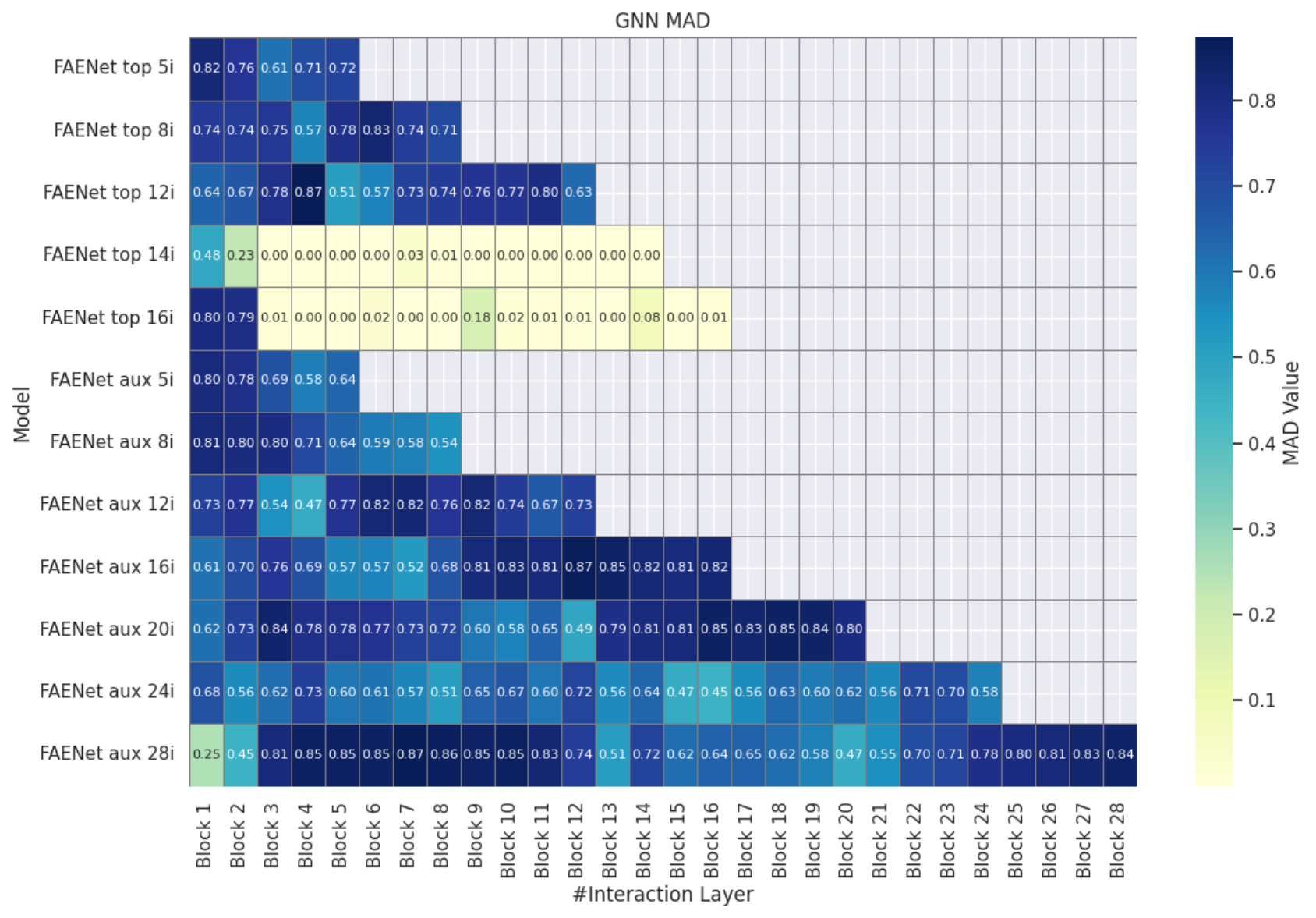}
    \caption{MAD values of the graph embeddings (averaged over 50 randomly sampled graphs of the train set) throughout the interaction layers for various models. ``FAENet top'' models are trained with the top configs of the classical FAENet model \cite{duval2023faenet} but with more epochs and lower batch size (128). ``FAENet aux'' models are our models trained on IS2RE with IS2RS auxiliary task. A model having xx interaction layers is indicated as ``XXi''.  
    }
    \label{fig:MAD_plot}
\end{figure}

To understand the reason for the performance drop for the classical FAENet IS2RE setup of \cite{duval2023faenet} when adding interaction layers in Table \ref{table:is2re_top_comp_interactions_fixed}, we plot the Mean Average Distance (MAD) \cite{MAD_chen}, averaged over 50 input adsorbate-catalyst pairs, of their embeddings throughout the interaction blocks in Figure \ref{fig:MAD_plot}. The classical FAENet models ("FAENet top") with 14 interaction layers or more see their latent node representations all collapse to almost the same vector, since the MAD goes to almost zero as we go deeper in the model's interaction layers, which is a manifestation of oversmoothing. 
Figure~\ref{fig:MAD_plot} shows that the models trained with Noisy Nodes IS2RS auxiliary task ("FAENet aux") do not suffer from oversmoothing (i.e. MAD going to zero) even when going as deep as 28 interaction layers.

Then, we compare the performances of our new FAENet models trained on IS2RE Noisy Nodes, varying the number of interaction layers, as summarized in Table~\ref{table:is2re_aux_mae_comp_interactions_5_28}. First, we observe a clear correlation between the performances and the number of blocks. Second, the best model trained with Noisy Nodes, which has 26 interaction blocks, outperforms the best models of \cite{duval2023faenet}, allowing the average MAE to decrease from 568 meV to 525 and the average EwT percentage to increase from 3.78\% to 4.43\%.
Details about our experimental setup are given in Appendix~\ref{Noisy Nodes appendix experimental setup}.
We observe that the throughput at inference is divided by two between the smallest (5 interaction layers) and biggest model (28 layers). The best tradeoff seems to be around 16 interaction layers, similar to what \cite{liao2023equiformer} choose for their models trained with IS2RS auxiliary task. 

The current most general state-of-the-art models for atomic property prediction, such as \cite{shoghi2024JMP} only make use of backbone GNNs (such as GemNet-OC \cite{gasteiger2022gemnetoc}) with not more than 6 interaction blocks.
Noisy Nodes proves to be a regularisation technique that can be used across models, having equivariant \cite{liao2023equiformer} or non-equivariant \cite{duval2023faenet} features, to leverage the power of more interaction layers and improve performance. On direct IS2RE, our experiments only show a slightly increasing performance with the number of blocks. Yet, unlocking the potential of deeper GNNs with this simple regularisation allows greater freedom in training or pretraining on larger datasets or for more complex tasks, such as relaxed IS2RE \ref{sec:relaxed_is2re}.
Our focus for FAENet was on IS2RE, but we expect similar results on S2EF, IS2RS, and other datasets such as QM9 or QM7-X, as obtained by \cite{godwin2022simple} with other GNNs such as GNS \cite{GNS-sanchez-gonzalez20a}.


\subsection{Learning equivariance with more interaction layers}
As previously stated, equivariant networks are preferred when handling 3D point clouds. However, in the field of protein structure prediction, we were caught by the AlphaFold3 architecture \cite{abramson2024accurate} that chose non-equivariant networks in its diffusion module, contrary to AlphaFold2 \cite{jumper2021highly}. Thus, we tested whether a deep GNN with no equivariance enforced could learn the equivariance from the data and match performances with canonicalized models using the FAENet backbone, as in Section \ref{sec:canonicalization}.

We compare our models trained on IS2RE with IS2RS auxiliary task in two settings: with SE(3)-SFA and without any canonicalization method (No-FA). The results are displayed in Table~\ref{table:is2re_aux_compFA_test_ri}.  
They reveal that, even with many additional interaction blocks (up to 26), both the MAE and the equivariance property do not improve compared to using SE(3)-SFA, meaning that imposing equivariance is still a beneficial inductive bias. Unfortunately, in our setting, having a deeper network does not allow to learn invariance and equivariance more effectively. Further analysis would be needed with larger datasets and other methods to reinforce this argument or find its limits (if any).

\subsection{Pre-training on different tasks}
With the release of larger datasets, the community has increasingly shifted towards pre-training and transfer learning approaches \cite{batatia2023foundation, shoghi2024JMP, deng2023chgnet}. We make a step in this direction in this subsection by pre-training our model on the S2EF, which contains roughly two orders of magnitude more data points, and by fine-tuning it on IS2RE, hoping to transfer some knowledge of atomic interactions. 
More precisely, we leverage the extensive S2EF dataset to utilize all trajectories while maintaining high inference throughput by performing direct energy predictions. Although training on S2EF is more time-consuming than on IS2RE, separating the learning of relaxation and molecular interactions may yield better results than training IS2RE directly from scratch.

Figure~\ref{fig:ft-is2re} in Appendix~\ref{sec:app_ft_is2re} shows that with this approach, the energy MAE starts at a better value during training but evolves much slower and converges to a slightly better result (0.53 eV vs. 0.55 eV on the validation ID split). This shows that while the model starts with a good representation of molecular dynamics, the differences are not significant because dynamics are not taken into account by the S2EF training process. Learning molecular interactions through auxiliary tasks with or without dynamics is helpful to achieve better performances, but this pre- or joint training needs to be correctly incorporated into the architecture so as not to overwrite the learned information during the downstream task. 
This opens the way for new architectures designed to leverage materials design knowledge between tasks and datasets~\citep{shoghi2024JMP}.
Whether by training with auxiliary tasks or on other datasets, there seems to be transfer learning and generalization capabilities in atomic property prediction, as in NLP. Hence, we recommend to further explore this promising area of research.

\section{Conclusion} 
In this study, we explored various techniques aimed at enhancing the performance of geometric GNNs for molecular modeling. Our empirical study covered several aspects, with the main observations being summarized below. 
\paragraph{Canonicalization methods.}
While exact methods provide the best theoretical guarantees for equivariance, approximative heuristics such as SFA seem to yield better performance. This opens questions about how to design canonicalizations that are the most effective in practice, beyond theoretical guarantees, and leaves a broader set of possibilities for the model design, too.

\paragraph{Graph creation.}
Although accurate graph construction is important, many viable options can be considered without significant differences in performance. 
Furthermore, physics-inspired modules such as Ewald-based message passing demonstrate improved performance for symmetry-constrained models such as SchNet but do not provide any benefits to more expressive models such as FAENet.

\paragraph{Auxiliary tasks.}
Implementing Noisy Nodes as an auxiliary task significantly enhances the performance of FAENet by leveraging the benefits of much deeper GNNs. As with pretraining on different tasks such as S2EF, there is evidence of transfer learning for atomic property prediction, and we recommend more exploration of this path in the flavor of \cite{shoghi2024JMP}.

Future work could focus on refining these techniques, exploring their applications across a wider spectrum of datasets, and developing new methods to combine the strengths of various approaches.

\section*{Acknowledgement}
The authors thank Sékou-Oumar Kaba, Derek Lim, Joshua David Robinson, and Hannah Lawrence for their insightful comments and discussions, as well as the anonymous reviewers for their suggestions and feedback. Supported in part by ANR (French National Research Agency) under the JCJC project GraphIA (ANR-20-CE23-0009-01) and the Canada CIFAR AI Chairs program. This research was enabled in part by computing resources provided by Mila (mila.quebec) and material support from NVIDIA Corporation in the form of computational resources.


\bibliography{main}
\bibliographystyle{icml2021}

\newpage
\appendix
\onecolumn

\section{Canonicalization} \label{appendix:cano}

\subsection{Stochastic Frame Averaging} \label{appendix:cano_sfa}
\paragraph{Frame Averaging}
We recall the idea of Frame Averaging introduced by \citet{puny2021frame}, in which $\mathcal{X}$ and $\mathcal{Y}$ denote normed linear space with respective representations $\rho_1$ and $\rho_2$ of a group $G$. In our case, the group of interest is $E(3)$. 
A \textit{frame} is defined as a function $\mathcal{F}: \mathcal{X} \to 2^{G}$ taking values in a non-empty subset of the group $G$ such that it is $G$-equivariant and bounded.
Under such conditions, and for every $\Phi: \mathcal{X} \to \mathcal{Y}$, the function $\langle \Phi \rangle_{\mathcal{F}}: \mathcal{X} \to \mathcal{Y}$, called \textit{average over the frame} $\mathcal{F}$, and defined as
\begin{equation}
{\langle \Phi \rangle}_{\mathcal{F}}: x \mapsto \frac{1}{|\mathcal{F}(x)|}\sum_{g\in\mathcal{F}(x)}\rho_2(g)\Phi(\rho_1(g)^{-1}x),
\end{equation}
is $G$-equivariant. This allows to create an arbitrary neural network, and guarantee the symmetry by averaging the outputs over a well-chosen frame.

\paragraph{Choosing the frame.} As in \citet{duval2023faenet}, the neural network takes as input $X$ the atoms position, $Z$ is their atomic numbers, and has outputs in $\mathcal{Y}$, where $\mathcal{Y} = \mathbb{R}^{n\times 3}$ in the case of force predictions, and $\mathcal{Y} = \mathbb{R}$ in the case of energy prediction. The group $E(3)$ only acts on $X$, and not $Z$.
A Principal Component Analysis (PCA) on the atomic structure allows to decompose the covariance matrix of the points cloud $\Sigma = U^T \Lambda U$ derived from the centroid of the positions $t = \frac{1}{n}X^T\bm{1}$, with $\Lambda$ the diagonal matrix containing the three eigenvalues $\lambda_1 > \lambda_2 > \lambda_3$ (assumed distinct because we consider non-planar structures). 
The frame is taken as 
\begin{equation}
\mathcal{F}(X) = \{(U, t) \mid U = [\pm u_1, \pm u_2, \pm u_3]\},
\end{equation}
which is a subset of $E(3)$. 
The authors prove that the frame $\mathcal{F}$ defined as such is $G$-equivariant and bounded.

\paragraph{Stochastic Frame Averaging.} This process requires to average the predictions of the neural network over $|\mathcal{F}(X)| = 2^3 =8$ elements of the frame. In order to make the computations faster, \citet{duval2023faenet} sample only one element from the frame instead of performing the average. 
Although Stochastic Frame-Averaging (SFA) does not have theoretical guarantees, it has been experimentally shown to learn almost perfect equivariance.

\subsection{SFA+SignNet} \label{annex:cano_signinv}
Initially proposed to help spectral graph representation learning, SignNet \cite{lim2023signinv} is a network which outputs are invariant to sign flips. The authors state that a continuous function $\eta : \R^n \to \R^d$ is \textit{sign-invariant} if and only if $\eta(x) = \kappa(x) + \kappa(-x)$ for some (freely chosen) continuous function $\kappa : \R^n \to \R^d$. 
$\text{SignNet} : \R^{n\times k} \to \R^{n\times k}$ is then defined as:
\begin{equation}
\text{SignNet}(x_1,\dots,x_k) = \mu \left([\kappa(x_i)+\kappa(-x_i)]_{i=1}^k \right),
\end{equation}
where $\mu$ and $\kappa$ are neural networks chosen freely.

We propose to apply such a network on the sampled element of the frame $U$ and parametrize $\mu$ and $\kappa$ either with MLPs or with VN-PointNets.
In order to constrain the output, we further orthonormalize the output with a Gram-Schmidt process:

\begin{equation}
U' = \text{Gram-Schmidt}(\text{SignNet}(U)).
\end{equation}

When parametrizing SignNet with MLPs, since we apply non-linearities on the orthogonal matrix $U$, there is no theoretical guarantee for the whole process to be $E(3)$-equivariant a priori. This is coherent with empirical observations when using this method, although training $\mu$ and $\kappa$ helps better enforce the equivariance.

When parametrizing SignNet with VN-PointNets, the whole network is made exactly $E(3)$-equivariant. As explained by \citet{lim2023signequiv} (section 2.2), the matrix $U$ is orthogonal and unique up to sign changes, while the SignNet function is both sign invariant by design and $O(3)$-equivariant thanks to the use of VN-PointNets.

\subsection{Vector Neurons Network} \label{appendix:cano_vnn}
VNNs are a class of $SO(3)$-equivariant models, where usual neurons are replaced with so-called Vector Neurons: for a given layer, non-linearities output a matrix of size $h \times 3$ instead of a vector of length $h$.

Using this framework, \citet{deng2021vector} re-implement classic operations such as linear layers, non-linearities, pooling operations, and normalization layers. They also prove that those layers are all $SO(3)$-equivariants, which allows to re-implement classical networks into their VN variant. In particular, they re-implement the VN variants of two well-known networks from the point clouds literature: PointNet \cite{qi2017pointnet} and DGCNN \cite{wang2019dgcnn}, and test them on classification, segmentation, and reconstruction tasks. They show that accuracy increases compared to the classic implementations and that their equivariance property is indeed (almost) perfectly enforced.

To obtain an $O(3)$ (and then $E(3)$) equivariance, the output of the VNN has to be further orthonormalized with a Gram-Schmidt process to canonicalize the representation in $O(3)$, as described by \citet{kaba2023equivariance}. 

The following Vector Neurons Networks (VNNs) are used:
\begin{itemize}
    \item VN-Pointnet with a varying number of VNLinearLeakyReLU layers (between 1 and 3), with the implementation of \citet{deng2021vector}.
    \item VN-DGCNN, with the implementation is the one of \citet{deng2021vector}.
\end{itemize}

To summarize, we use VNNs to learn the transformation $U$ from the positions $X$:
\begin{align*}
    \text{VNN} : \R^{n\times 3}& \to \R^{3\times 3} \\
    X& \mapsto U.
\end{align*}
$U$ is then orthonormalized with a Gram-Schmidt process.

\subsection{Experimental comparisons} \label{annex:cano_exp}
OC20 IS2RE: Tables \ref{table:lcf_is2re_acc} and \ref{table:lcf_is2re_perf}

OC20 S2EF: Tables \ref{table:lcf_s2ef_acc} and \ref{table:lcf_s2ef_perf}


QM9: Table \ref{table:lcf_qm9_acc}


In all OC20 experiments of this section, for a fair assessment, SFA is used in 3D mode, i.e. without the computational trick to force the $z$ axis to remain fixed during canonicalization, which is specific to OC20. The same goes for the methods derived from SFA. As a consequence, the reported performances are lower than reported in other sections.

\begin{table*}[htb]
\centering
\resizebox{0.8\textwidth}{!}{
    \begin{tabular}{l|c|cccccc}
    Canonicalization & Cano. trainable & 2D Rotation $\downarrow$ & 3D Rotation $\downarrow$ & Reflection $\downarrow$ \\ 
    & parameters &   Invariance & Invariance  & Invariance \\
    \hline
    SFA & 0 & $1.29 \cdot 10^{-2}$ & $1.32 \cdot 10^{-2}$ & $1.30 \cdot 10^{-2}$ \\ 
    Untrained SFA+MLP-SignNet & 0 & $1.01 \cdot 10^{-1}$ & $1.00 \cdot 10^{-1}$ & $9.71 \cdot 10^{-2}$\\
    Trained SFA+MLP-SignNet & 454 & $4.21 \cdot 10^{-2}$ & $7.74 \cdot 10^{-2}$ & $4.00 \cdot 10^{-2}$ \\
    Untrained SFA+VN-SignNet & 0 & $6.89 \cdot 10^{-3}$ & $7.58 \cdot 10^{-3}$ & $7.37 \cdot 10^{-3}$\\
    Trained SFA+VN-SignNet & 2,620 & $2.45 \cdot 10^{-2}$ & $2.66 \cdot 10^{-2}$ & $2.43 \cdot 10^{-2}$ \\
    \hline
    Untrained VN-Pointnet (2 hid.) & 0  & $4.61 \cdot 10^{-3}$ & $4.62 \cdot 10^{-3}$ & $4.62 \cdot 10^{-3}$ \\ 
    Trained VN-Pointnet (2 hid.) & 1,310 & $\mathbf{3.63 \cdot 10^{-3}}$ & $\mathbf{3.72 \cdot 10^{-3}}$ & $\mathbf{3.80 \cdot 10^{-3}}$ \\ 
    Untrained VN-Pointnet (1 hid.) & 0 & $4.28 \cdot 10^{-3}$ & $4.28 \cdot 10^{-3}$ & $4.29 \cdot 10^{-3}$\\ 
    Untrained VN-Pointnet (0 hid.) & 0 & $2.76 \cdot 10^{-2}$ & $2.76 \cdot 10^{-2}$ & $2.79 \cdot 10^{-2}$\\ 
    Trained VN-Pointnet (0 hid.) & 24 & $1.86 \cdot 10^{-2}$ & $2.31 \cdot 10^{-2}$ & $2.36 \cdot 10^{-2}$ \\ 
    Untrained VN-DGCNN & 0 & $3.03 \cdot 10^{-2}$ & $3.08 \cdot 10^{-2}$ & $3.11 \cdot 10^{-2}$ \\ 
    Trained VN-DGCNN & 663,804 & $9.89 \cdot 10^{-3}$ & $2.49 \cdot 10^{-2}$ & $9.10 \cdot 10^{-3}$ \\ 
    \end{tabular}
}
\caption{Invariance comparison of canonicalization methods on OC20 IS2RE dataset. The FAENet backbone for this task and dataset has 4,147,731 parameters (5 interaction blocks). We measure the rotation invariance and reflection invariance property as the difference in prediction between every samples D1 (of the ID val split) and D2 defined as a SO(3)
transformation of D1, in eV.}
\label{table:lcf_is2re_acc}
\end{table*}

\clearpage 
\begin{table*}[htb]
\centering
\resizebox{1\textwidth}{!}{
    \begin{tabular}{l|cc|cc|cc|cc}
    & \multicolumn{2}{c|}{ID} & \multicolumn{2}{c|}{OOD-CAT} & \multicolumn{2}{c|}{OOD-ADS} & \multicolumn{2}{c}{OOD-BOTH} \\
    Canonicalization & EwT (\%) $\uparrow$& MAE (meV) $\downarrow$& EwT (\%) $\uparrow$& MAE (meV) $\downarrow$ & EwT (\%) $\uparrow$ & MAE (meV) $\downarrow$& EwT (\%) $\uparrow$& MAE (meV) $\downarrow$\\
    \hline
    SFA & 4.40 &  566 &  4.12&  563 &  2.56&  652&  2.77 & 594  \\ 
    Untrained SFA+MLP-SignNet & 4.48 & \textbf{554} & 4.46 & 552 & 2.75 & \textbf{637} & 2.88 & \textbf{576}\\
    Trained SFA+MLP-SignNet & 4.46 & \textbf{554} & 4.51 & \textbf{551} & 2.67 & 642 & 2.78 & 586\\
    Untrained SFA+VN-SignNet & \textbf{4.69} & 563 & 4.60 & 558 & 2.62 & 651 & 2.59 & 595 \\
    Trained SFA+VN-SignNet & 4.25 & 572 & 4.27 & 568 & \textbf{2.92} & 658 & \textbf{2.97} & 596 \\
    \hline
    Untrained VN-Pointnet (2 hid.) & 4.09 & 567 & \textbf{4.66} & 565 & 2.60 & 673 & 2.85 & 615 \\ 
    Trained VN-Pointnet (2 hid.) & 4.12 & 568 & 4.33 & 563 & 2.77 & 658 & 2.75 & 604 \\ 
    Untrained VN-Pointnet (1 hid.) & 4.37 & 565 & 4.20 & 561 & 2.64 & 666 & 2.73 & 614\\
    Untrained VN-Pointnet (0 hid.) & 4.01 & 581 & 3.92 & 571 & 2.75 & 660 & 2.64 & 615\\
    Trained VN-Pointnet (0 hid.) & 4.14 & 567 & 4.36 & 563 & 2.56 & 675 & 2.88 & 614\\
    Untrained VN-DGCNN & 4.31 & 567 & 4.14 & 562 & 2.58 & 660 & 2.72 & 610 \\ 
    Trained VN-DGCNN & 4.42 & 560 & 4.40 & 556 & 2.78 & 656 & 2.81 & 601 \\ 
    \end{tabular}
    }
\caption{Performance comparison of canonicalization methods on OC20 IS2RE dataset. All models were trained for 12 epochs using \citet{duval2023faenet} config.}
\label{table:lcf_is2re_perf}
\end{table*}

\begin{table*}[htb]
\centering
\resizebox{0.9\textwidth}{!}{
    \begin{tabular}{l|c|cc|cc} & Cano. trainable & \multicolumn{2}{c|}{Energy invariance} & \multicolumn{2}{c}{Forces equivariance} \\
    Canonicalization & parameters & 3D Rotation $\downarrow$ & Reflection $\downarrow$ & 3D Rotation $\downarrow$ & Reflection $\downarrow$ \\ 
    \hline
    SFA & 0 & $1.88 \cdot 10^{-2}$ & $1.88 \cdot 10^{-2}$ & $7.17 \cdot 10^{-2}$ & $8.34 \cdot 10^{-3}$ \\
    Untrained SFA+MLP-SignNet & 0 & $7.81 \cdot 10^{-2}$ & $7.61 \cdot 10^{-2}$ & $7.44 \cdot 10^{-2}$ & $2.04 \cdot 10^{-2}$\\
    Trained SFA+MLP-SignNet & 454 & $6.57 \cdot 10^{-2}$ & $3.57 \cdot 10^{-2}$ & $7.35 \cdot 10^{-2}$ & $1.17 \cdot 10^{-2}$ \\
    Untrained SFA+VN-SignNet & 0 & $2.07 \cdot 10^{-2}$ & $2.04 \cdot 10^{-2}$ & $6.86 \cdot 10^{-2}$ & $9.42 \cdot 10^{-3}$\\
    Trained SFA+VN-SignNet & 2,620 & $1.92 \cdot 10^{-2}$ & $1.89 \cdot 10^{-2}$ & $6.55 \cdot 10^{-2}$ & $8.72 \cdot 10^{-3}$ \\
    \hline
    Untrained VN-Pointnet (2 hid.) & 0 & $1.80 \cdot 10^{-2}$ & $1.80 \cdot 10^{-2}$ & $6.92 \cdot 10^{-2}$ & $8.78 \cdot 10^{-3}$ \\ 
    Trained VN-Pointnet (2 hid.) & 1,310 & $1.67 \cdot 10^{-2}$ & $1.67 \cdot 10^{-2}$ & $6.89 \cdot 10^{-2}$ & $8.77 \cdot 10^{-3}$ \\
    Untrained VN-Pointnet (0 hid.) & 0 & $3.50 \cdot 10^{-2}$ & $3.49 \cdot 10^{-2}$ & $6.96 \cdot 10^{-2}$ & $1.08 \cdot 10^{-2}$ \\
    Trained VN-Pointnet (0 hid.) & 24 & $3.31 \cdot 10^{-2}$ & $3.34 \cdot 10^{-2}$ & $7.00 \cdot 10^{-2}$ & $1.05 \cdot 10^{-2}$ \\ 
    Untrained VN-DGCNN & 0 & $\mathbf{1.50 \cdot 10^{-2}}$ & $\mathbf{1.50 \cdot 10^{-2}}$ & $\mathbf{6.83 \cdot 10^{-2}}$ & $\mathbf{3.58 \cdot 10^{-3}}$\\
    Trained VN-DGCNN & 663,804 & $2.02 \cdot 10^{-2}$ & $1.50 \cdot 10^{-2}$ & $6.91 \cdot 10^{-2}$ & $8.09 \cdot 10^{-3}$ \\
    \end{tabular}
}
\caption{Equivariance comparison of canonicalization methods on OC20 S2EF dataset. The FAENet backbone for this task and dataset has 5,675,410 parameters (7 interaction blocks). We measure the energy rotation invariance, energy reflection invariance, force rotation equivariance, and force reflection equivariance properties as the difference in prediction between every sample D1 (of the ID val split) and D2 defined as a SO(3) transformation of D1, in eV.}
\label{table:lcf_s2ef_acc}
\end{table*}

\begin{table*}[htb]
\centering
\resizebox{0.95\textwidth}{!}{
    \begin{tabular}{l|cccc|cccc}
    & \multicolumn{4}{c|}{Energy MAE (mEV) $\downarrow$} & \multicolumn{4}{c}{Force MAE (mEV) $\downarrow$}\\
    Canonicalization & ID & OOD Cat & OOD Ads & OOD Both & ID & OOD Cat & OOD Ads & OOD Both \\
    \hline
    SFA & 424 & 445 & 579 & 680 & 55.6 & 55.2 & 63.2 & 74.6  \\
    Untrained SFA+MLP-SignNet & \textbf{420} & \textbf{444} & \textbf{515} & \textbf{631} & \textbf{54.0} & \textbf{53.8} & \textbf{61.4} & \textbf{72.4} \\
    Trained SFA+MLP-SignNet & 422 & 446 & 558 & 666 & 54.2 & 53.9 & 62.7 & 73.8 \\
    Untrained SFA+VN-SignNet & 439 & 458 & 565 & 673 & 56.5 & 56.0 & 65.3 & 76.7 \\
    Trained SFA+VN-SignNet & 442 & 464 & 590 & 701 & 58.0 & 57.5 & 65.2 & 76.9 \\
    \hline
    Untrained VN-Pointnet (2 hid.) & 435 & 455 & 596 & 697 & 56.0 & 55.6 & 66.9 & 77.5 \\ 
    Trained VN-Pointnet (2 hid.) & 435 & 453 & 585 & 696 & 56.1 & 55.8 & 64.2 & 75.6 \\
    Untrained VN-Pointnet (0 hid.) & 440 & 459 & 597 & 705 & 56.0 & 55.6 & 64.7 & 75.8   \\
    Trained VN-Pointnet (0 hid.) & 440 & 459 & 572 & 671 & 55.8 & 55.4 & 63.7 & 74.9 \\ 
    Untrained VN-DGCNN & 456 & 474 & 593 & 763 & 55.7 & 55.5 & 65.8 & 76.9 \\
    Trained VN-DGCNN & 432 & 453 & 662 & 762 & 55.5 & 55.2 & 71.2 & 80.7 \\
    \end{tabular}
    }
\caption{Performance comparison of canonicalization methods on OC20 S2EF dataset. All models were trained for 12 epochs using \citet{duval2023faenet} config.}
\label{table:lcf_s2ef_perf}
\end{table*}

\begin{table*}[htb]
\centering
\resizebox{0.8\textwidth}{!}{
    \begin{tabular}{l|c|cc|cc} & Cano. trainable & \multicolumn{2}{c|}{MAE (meV) $\downarrow$} & \multicolumn{2}{c}{Energy invariance (eV)} \\
    Canonicalization & parameters & ID & Test & 3D Rotation $\downarrow$ & Reflection $\downarrow$ \\ 
    \hline
    SFA & 0 & \textbf{9.20} & \textbf{9.06} &  $1.65 \cdot 10^{-3}$ & $1.76 \cdot 10^{-3}$ \\
    Untrained SFA+MLP-SignNet & 0 & 11.3 & 11.2 & $2.12 \cdot 10^{-3}$ & $2.20 \cdot 10^{-3}$  \\
    Trained SFA+MLP-SignNet & 454 & 10.5 & 10.7 & $1.60 \cdot 10^{-3}$ & $1.66 \cdot 10^{-3}$ \\
    Untrained SFA+VN-SignNet & 0 & 9.41 & 9.40 & $1.28 \cdot 10^{-3}$ & $1.37 \cdot 10^{-3}$  \\
    Trained SFA+VN-SignNet & 2,620 & 10.1 & 10.2 & $1.33\cdot 10^{-3}$ & $1.40 \cdot 10^{-3}$ \\
    \hline
    Untrained VN-Pointnet (2 hid.) & 0 & 10.4 & 10.3 & $\mathbf{1.19} \cdot 10^{-3}$ & $1.30 \cdot 10^{-3}$  \\ 
    Trained VN-Pointnet (2 hid.) & 1,310 & 10.1 & 9.85 & $1.30 \cdot 10^{-3}$ & $1.44 \cdot 10^{-3}$\\
    Untrained VN-Pointnet (0 hid.) & 0 & 9.51 & 9.49 & $1.21 \cdot 10^{-3}$ & $\mathbf{1.24} \cdot 10^{-3}$  \\
    Trained VN-Pointnet (0 hid.) & 24 & 11.4 & 11.5 & $1.64 \cdot 10^{-3}$ & $1.74 \cdot 10^{-3}$\\ 
    Untrained VN-DGCNN & 0 & 9.92 & 9.94 & $1.32 \cdot 10^{-3}$ & $1.46 \cdot 10^{-3}$  \\
    Trained VN-DGCNN & 663,804 & 9.34 & 9.25 & $1.79 \cdot 10^{-3}$ & $1.79 \cdot 10^{-3}$ \\ 
    \end{tabular}
}
\caption{Equivariance and performance comparison of canonicalization methods on QM9 dataset for the target property $U_0$ (internal energy at 0 Kelvin). The FAENet backbone for this task has 6,495,127 parameters (5 interaction blocks). We measure the energy rotation invariance as the difference in prediction between every samples D1 (of the ID val split) and D2 defined as a SO(3) transformation of D1, in eV. All models were trained for 300 epochs using \citet{duval2023faenet} config.}
\label{table:lcf_qm9_acc}
\end{table*}

\newpage
\subsection{Relaxations from S2EF model for IS2RE}
\label{sec:app_relax}
The models used to run the experiments with the relaxation methods were trained on the 2M train split of the S2EF dataset from OC20. This dataset has been shown to converge to similar performances as the complete dataset, which is way larger and takes too long to train on~\citep{gasteiger2022gemnetoc}. We report in Table~\ref{table:relaxations_s2ef} the performances of these models, which might help interpret some of the results for the relaxation.

\begin{table*}[ht]
\centering
\resizebox{0.6\textwidth}{!}{
    \begin{tabular}{l|cccc|cccc|cccc}
    Model & \multicolumn{1}{c}{EwT $\downarrow$} & \multicolumn{1}{c}{Force MAE $\downarrow$} & \multicolumn{1}{c}{Forces $\cos\quad$$\uparrow$} \\
    \hline
    FAENet with SFA & 10.7 & 0.044 & 0.32 \\
    FAENet with Untrained PointNet & 10.5 & 0.0043 & 0.33 \\
    FAENet without SFA & 10.0 & 0.042 & 0.34\\
    SchNet Base& 5.1 &  0.061 & 0.07 \\
    \end{tabular}
    }
\caption{Performance comparison of models on OC20 S2EF dataset on the VAL-ID split. The energy within threshold, Force MAE, and $\cos$ similarity are reported for these S2EF models that are then used for relaxations. Note that this method yields way longer training and inference times when compared to direct IS2RE as reported.}
\label{table:relaxations_s2ef}
\end{table*}


\section{Graph creation study}

\subsection{Cutoff}
\label{sec:app_cutoff}
The cutoff defines the distance within which a link between two atoms is created. All atoms that are at a distance smaller than this cutoff will be linked. However, in order to avoid cluttering the created graphs, most methods impose a maximum number of neighbors for every atom. This parameter is usually taken around 40 neighbors.

\begin{table}[ht]
\centering
\resizebox{\textwidth}{!}{
\begin{tabular}{l|cc|cc|cc|cc}
& \multicolumn{2}{c|}{ID} & \multicolumn{2}{c|}{OOD-ADS} & \multicolumn{2}{c|}{OOD-CAT} & \multicolumn{2}{c}{OOD-BOTH} \\
Model & EwT (\%) $\uparrow$& MAE (eV) $\downarrow$& EwT (\%) $\uparrow$& MAE (eV) $\downarrow$ & EwT (\%) $\uparrow$ & MAE (eV) $\downarrow$& EwT (\%) $\uparrow$& MAE (eV) $\downarrow$\\
\hline
Cutoff 30 - Max neighbours 40 & 2.65 & 0.697 & 1.45 & 0.906 & 2.86 & 0.691 & 1.53 & 0.846 \\
Cutoff 20 - Max. neighbours 40 & 3.08 & 0.673 & 1.85 & 0.808 & 2.86 & 0.669 & 1.86 & 0.757 \\
Cutoff 20 - Max. neighbours 10 & 2.25 & 0.768 & 1.51 & 0.988 & 2.52 & 0.754 & 1.38 & 0.928 \\
Cutoff 10 - Max. neighbours 50 & 4.17 & 0.553 & 2.81 & 0.640 & 4.12 & 0.551 & 3.02 & 0.585 \\
Cutoff 10 - Max. neighbours 40 & 4.29 & 0.555 & 2.95 & 0.631 & 4.33 & 0.553 & 2.71 & 0.587 \\
Cutoff 10 - Max. neighbours 30 & 4.43 & 0.551 & 2.65 & 0.655 & 4.51 & 0.552 & 2.51 & 0.611 \\
Cutoff 10 - Max. neighbours 20 & 4.38 & 0.551 & 2.46 & 0.676 & 4.45 & 0.551 & 2.55 & 0.621 \\
Cutoff 10 - Max. neighbours 10 & 4.49 & 0.553 & 2.84 & 0.627 & 4.34 & 0.549 & 3.01 & 0.582 \\
Cutoff 6 - Max. neighbours 40 & 4.31 & 0.553 & 3.00 & 0.626 & 4.39 & 0.554 & 2.81 & 0.577 \\
Cutoff 1 - Max. neighbours 40 & 1.35 & 1.069 & 1.32 & 1.112 & 1.33 & 1.051 & 1.37 & 1.018 \\
\end{tabular}
}
\caption{Impact of the cutoff on the performances of FAENet on the OC20 IS2RE task. Full table on all validation splits.}
\label{table:cutoff_long}
\end{table}

\subsection{Ewald-based Long-Range Message Passing}
\label{sec:app_ewald}
\begin{figure}[H]
  \centering
  \includegraphics[width=0.6\linewidth]{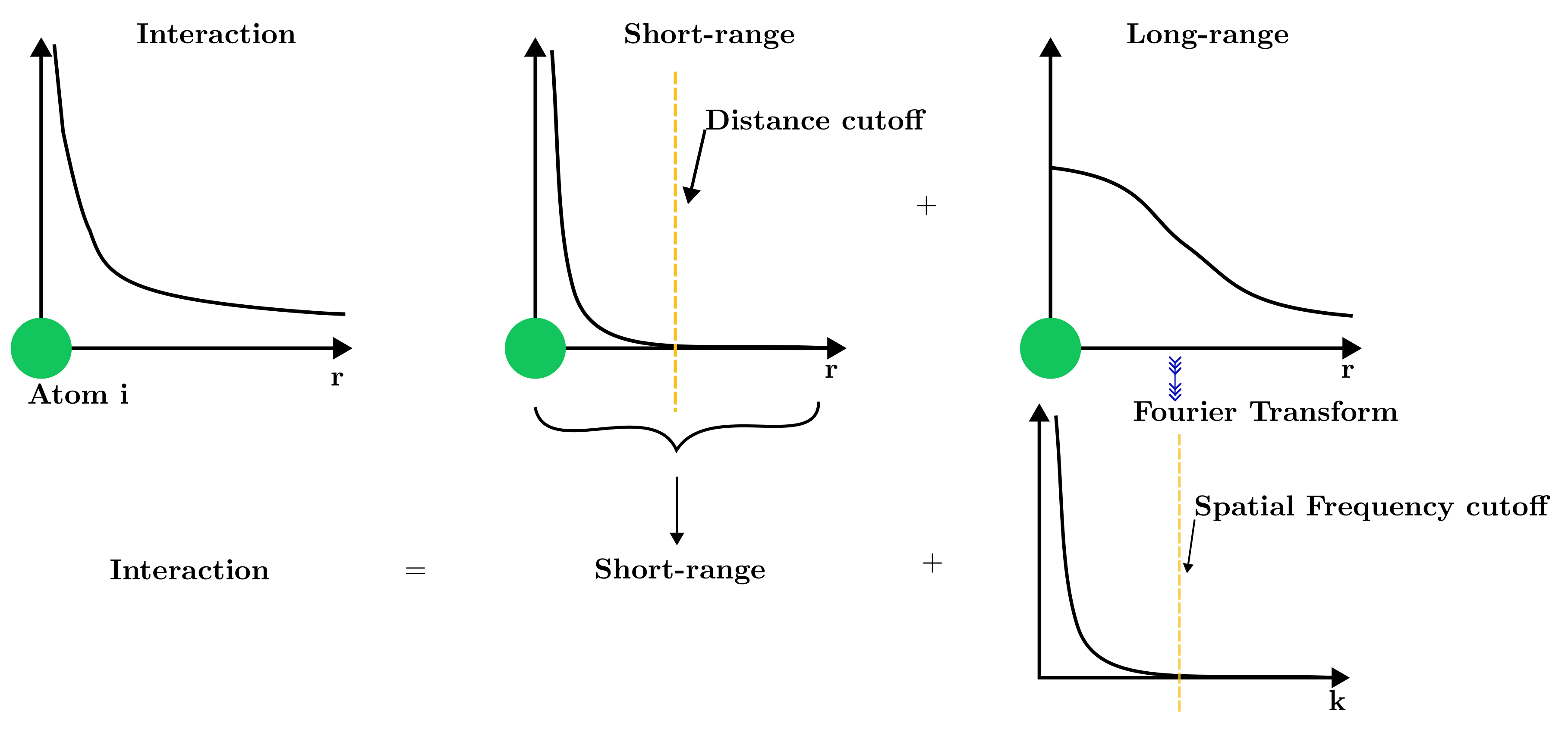}
  \caption{Ewald Summation interactions. The interaction term (left) is the result of the short-range interaction (middle) and the long-term interaction (right) which are both computed using cutoffs on respectively the real and the Fourier space. Adapted from \cite{kosmala2023ewald}}
  \label{fig:ewald_summation}
\end{figure}

The main idea behind Ewald summation used in section~\ref{sec:ewald} is to decompose the electrostatic interaction potential with a given charge into a short-range interaction and a long-range interaction term. The short-range contribution can be computed with real spatial features and the long-range contribution is computed using a Fourier transform. This principle is illustrated in Figure \ref{fig:ewald_summation}. This allows for computational methods in electrostatics to converge faster and with higher accuracy because the long-range interaction becomes more tractable.
In the case of GNNs, the short-range interaction is already computed by the currently implemented interaction blocks using a distance cutoff between the atoms, which omits the negligible parts of this interaction for further atoms. However, the heavy-tail of long-range interactions is then reported to a new term in real space, which doesn't diverge anymore for closer atoms. It can then be computed using the same cutoff idea but in the Fourier space where a \textit{spatial frequency cutoff} is used to make it tractable as introduced in \citet{kosmala2023ewald}. 

\paragraph{Periodic case.} In the case where there exists a spatial periodic tiling of materials (OC20 for example), it is possible to define the set of periodic cells localization $\Lambda = \{\lambda_1\mathbf{v_1}, \lambda_2\mathbf{v_2}, \lambda_3\mathbf{v_3}\mid (\lambda_1, \lambda_2, \lambda_3)\in\mathbb{Z}^3\}$, where $\mathbf{v_1}, \mathbf{v_2}, \mathbf{v_3}$ define the periodic cell lattice, similarly to the periodic interval in the 1D case. In the real space, the long-range interaction component would be written as a sum over all of the elements on the infinite tiling, which can be decomposed as a Fourier series expansion using the reciprocal lattice $\Lambda'$. This reciprocal lattice would be similar to the $2\pi$ in the 1D case of the Fourier transform. It is the periodic space of all the wavevectors of the Fourier series. This results in the proposed expansion for Ewald message passing:
\begin{equation}
  M^{lr}(x_i) = \sum_{\mathbf{k}\in\Lambda'}\exp{(i\mathbf{k}^Tx_i)}\cdot \sum_{j\in \mathcal{S}} h_j \exp{(-i\mathbf{k}^Tx_j)} \cdot \hat{\Phi}^{lr}(||\mathbf{k}||),  
\end{equation}
where $M^{lr}(x_i)$ corresponds to the long-range message computed at node $i$ from all of the nodes in the system $S$, and $\hat{\Phi}^{lr}$ is a learned Fourier coefficient of a radial basis function representing the interaction. The cutoff in the Fourier basis $c_k$ is then set to make the sum finite over the set $\{k\in \Lambda', ||k|| \leq c_k\}$. Since the number of wavevectors used for the computation is finite, $\hat{\Phi}(\|\mathbf{k}\|)$ is learned for every $\mathbf{k}$.
Since Ewald summation applies to periodic structures, the authors of Ewald message passing \citet{kosmala2023ewald} propose tricks to deal with the aperiodic case by assuming an infinite tiling.



\begin{figure}[H]
    \centering
    \subfloat[SchNet with Ewald - Interaction Blocks]{\includegraphics[width=0.48\linewidth]{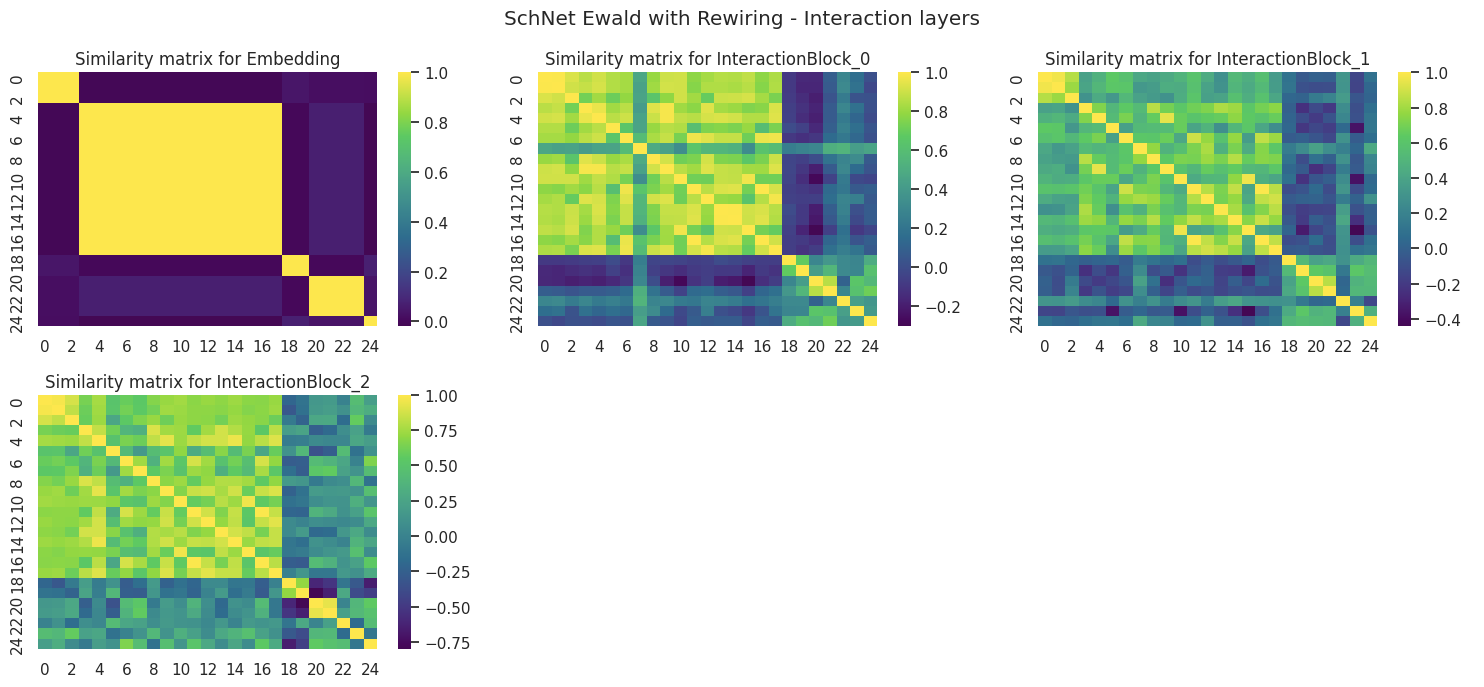}}
    \hfill
    \subfloat[FAENet with Ewald - Interaction Blocks]{\includegraphics[width=0.48\linewidth]{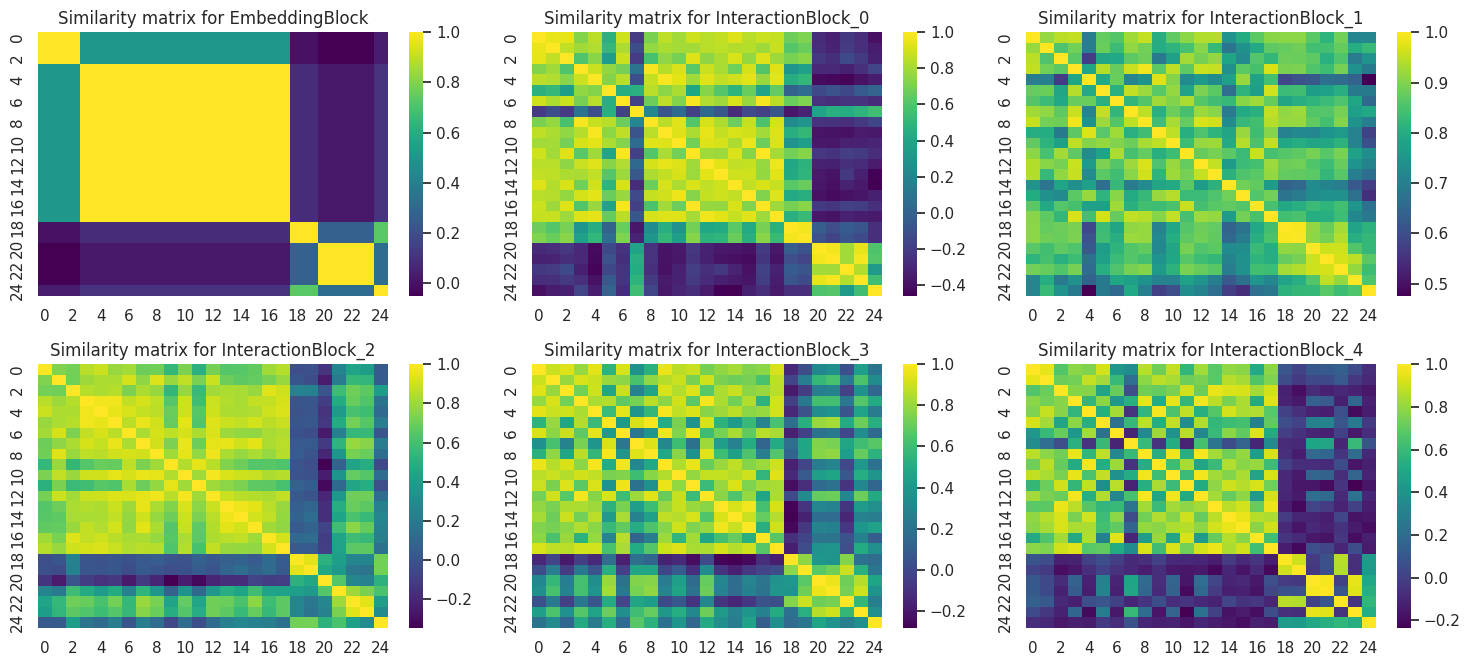}} \\
    \subfloat[SchNet with Ewald - Ewald Blocks]{\includegraphics[width=0.48\linewidth]{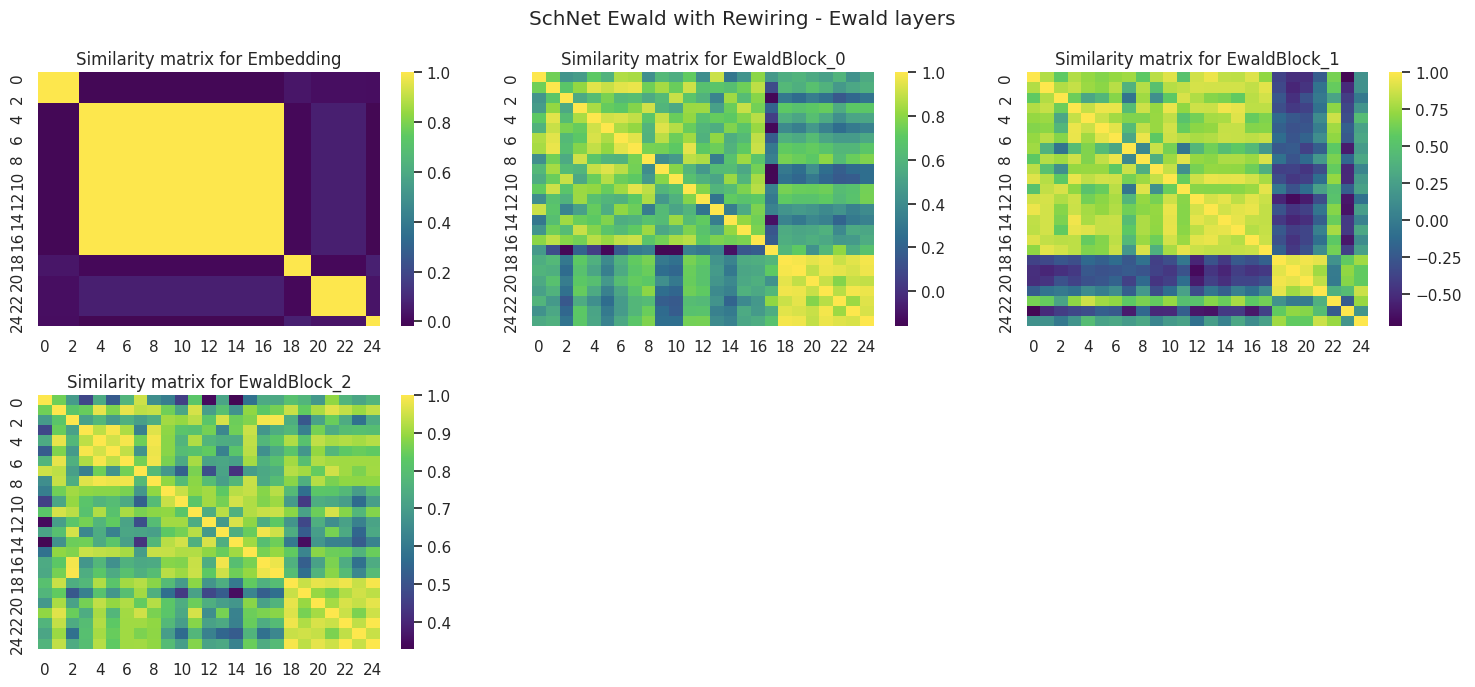}}
    \hfill
    \subfloat[FAENet with Ewald - Ewald Blocks]{\includegraphics[width=0.48\linewidth]{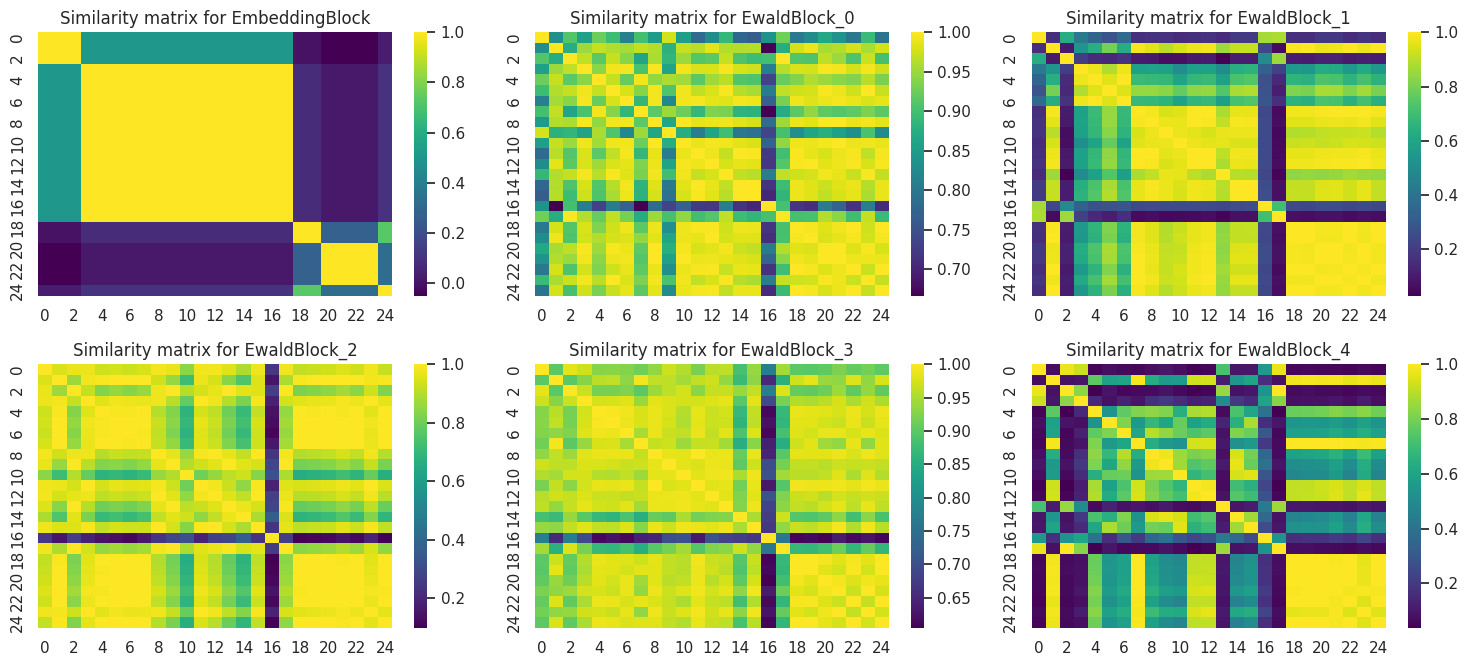}}
    \caption{Similarity matrix of the embeddings of the atoms of a system for different interaction blocks on FAENet and SchNet with Ewald message passing on the models. The visualized layers here are the standard interaction blocks in the first row and the Ewald interaction blocks in the second row. The two outputs are summed to get the final representation for Ewald shown in  Figure~\ref{fig:ewald_embeddings}.}
    \label{fig:ewald_embeddings_diverse}
\end{figure}

\begin{figure}[H]
    \centering
    \subfloat[SchNet without Ewald]{\includegraphics[width=0.48\linewidth]{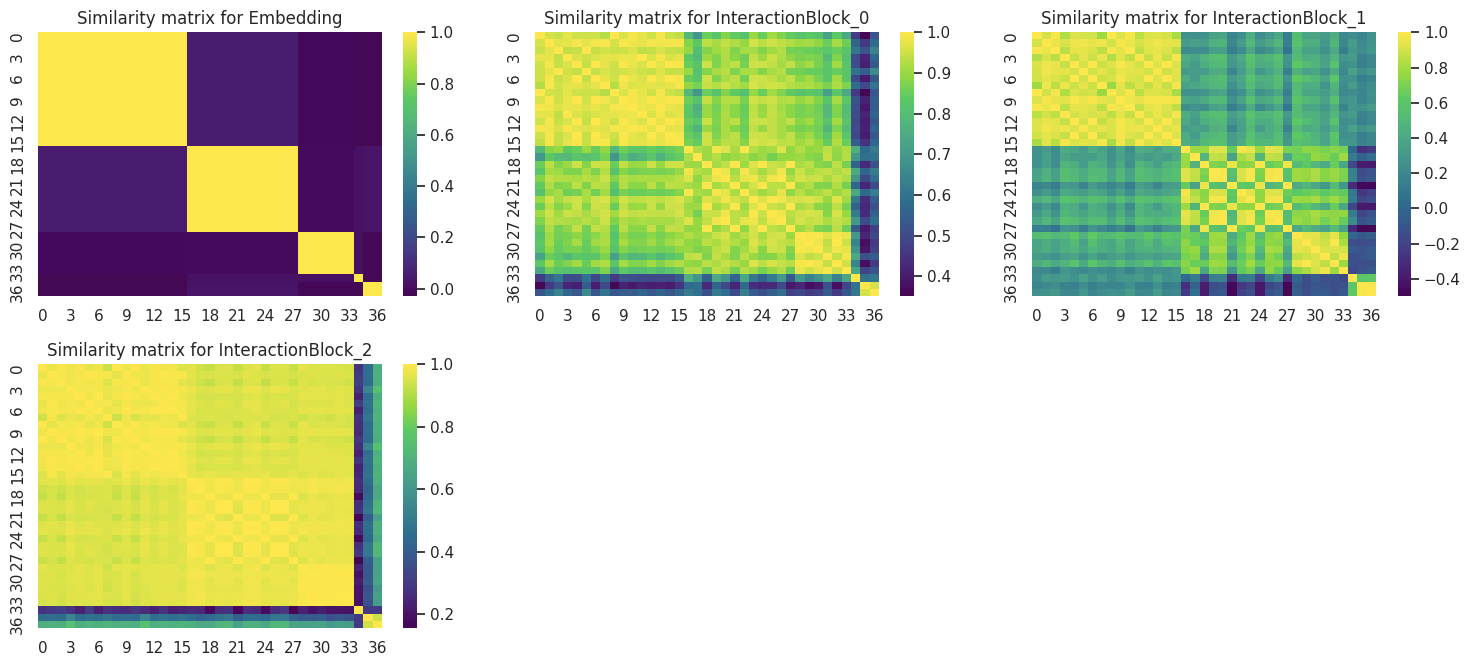}}
    \hfill
    \subfloat[FAENet without Ewald]{\includegraphics[width=0.48\linewidth]{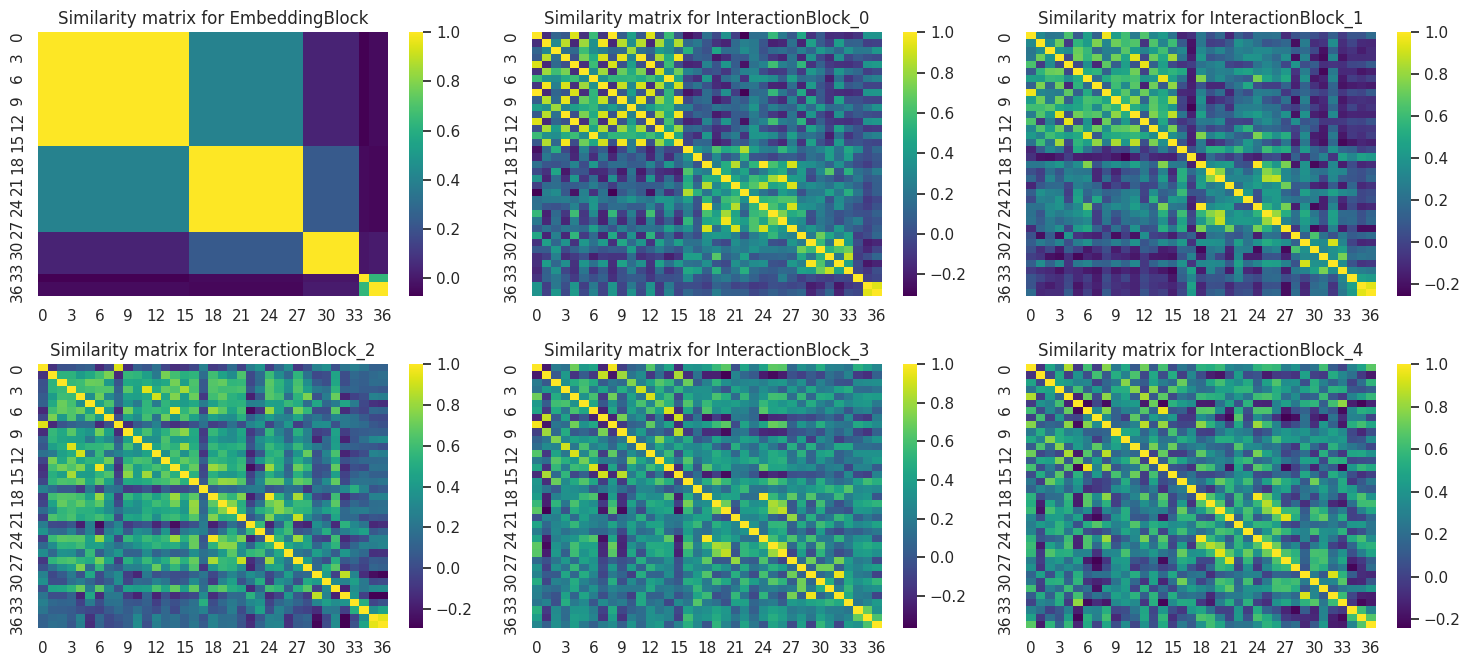}}\\
    \subfloat[SchNet with Ewald]{\includegraphics[width=0.48\linewidth]{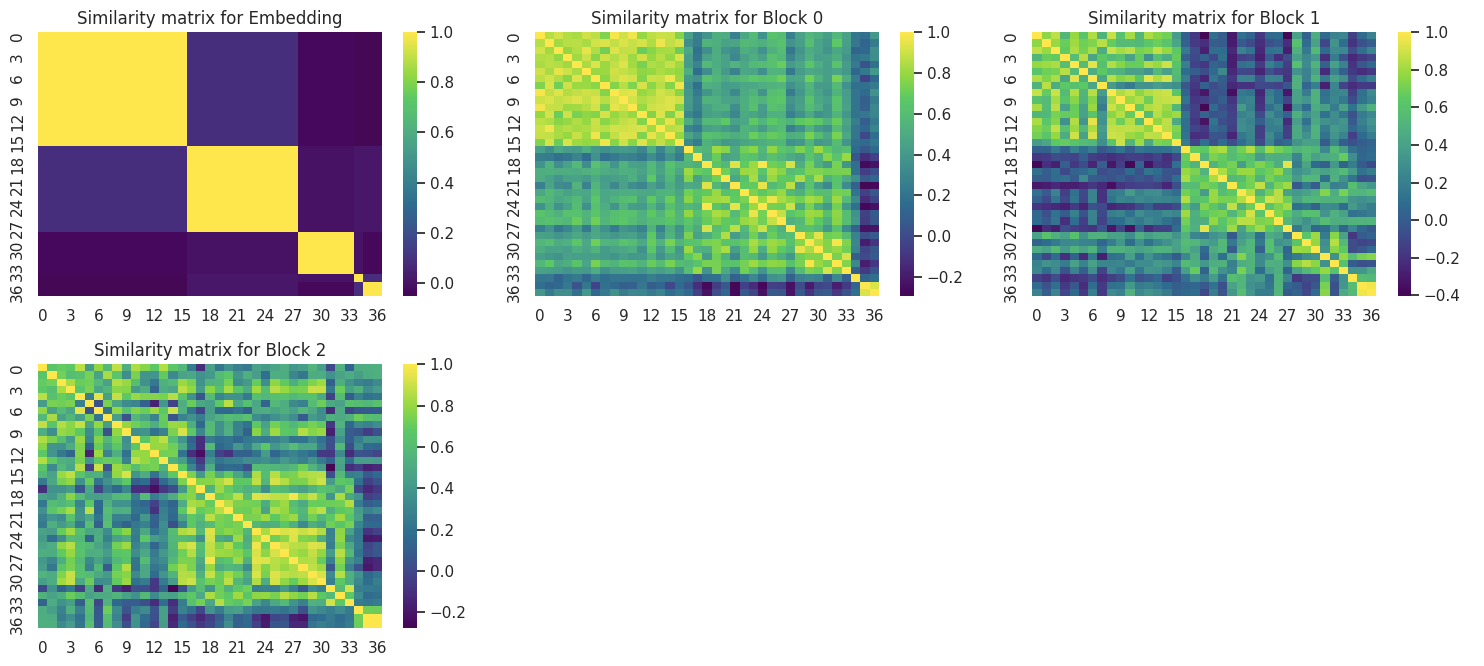}}
    \hfill
    \subfloat[FAENet with Ewald]{\includegraphics[width=0.48\linewidth]{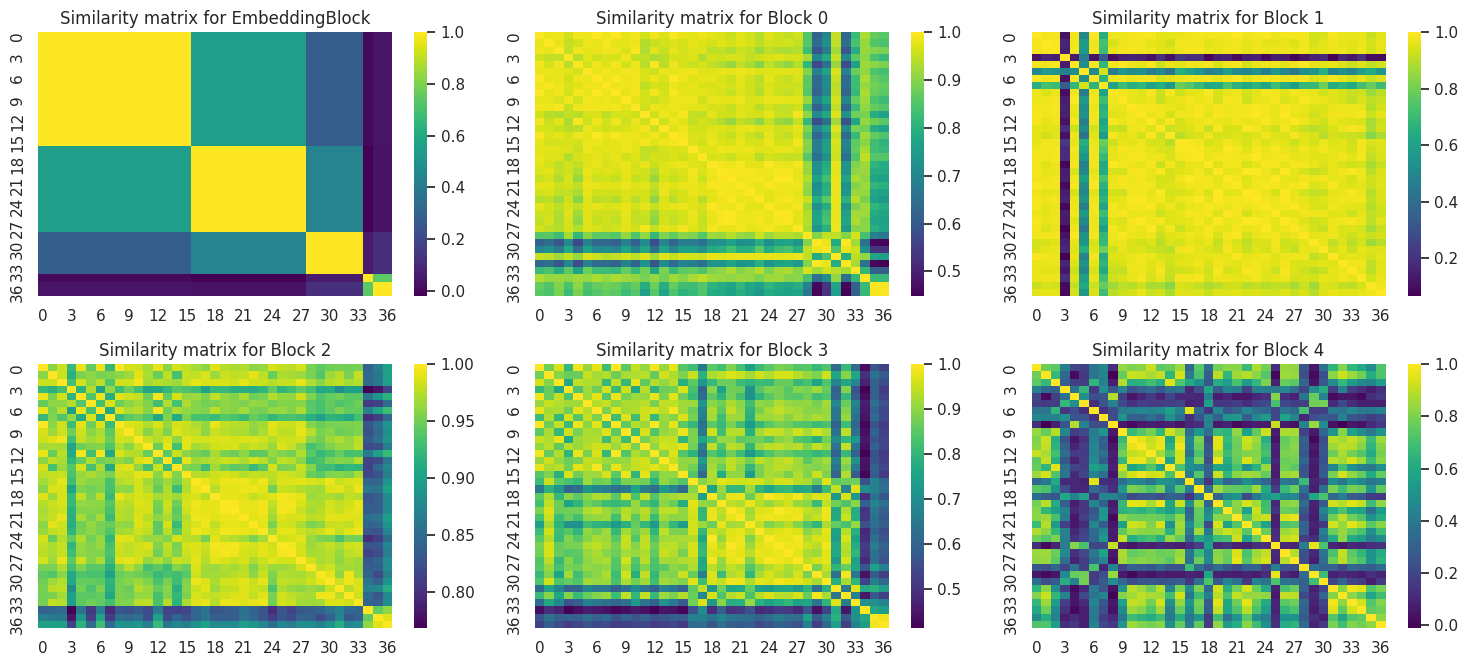}}
    \caption{Same plots as Figure~\ref{fig:ewald_embeddings} but with a second randomly picked system from the OC20 train split.}
    \label{fig:ewald_embeddings2}
\end{figure}

\begin{figure}[H]
    \centering
    \subfloat[SchNet with Ewald - Interaction Blocks]{\includegraphics[width=0.48\linewidth]{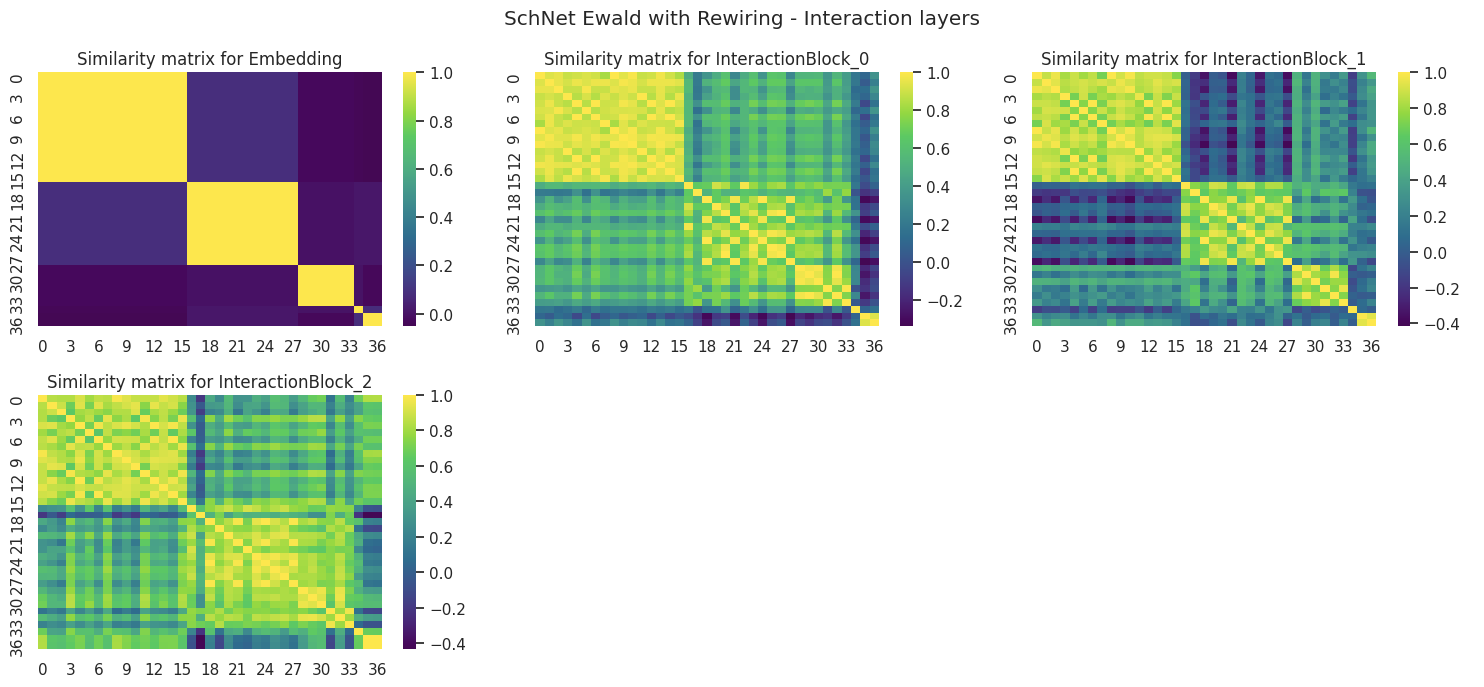}}
    \hfill
    \subfloat[FAENet with Ewald - Interaction Blocks]{\includegraphics[width=0.48\linewidth]{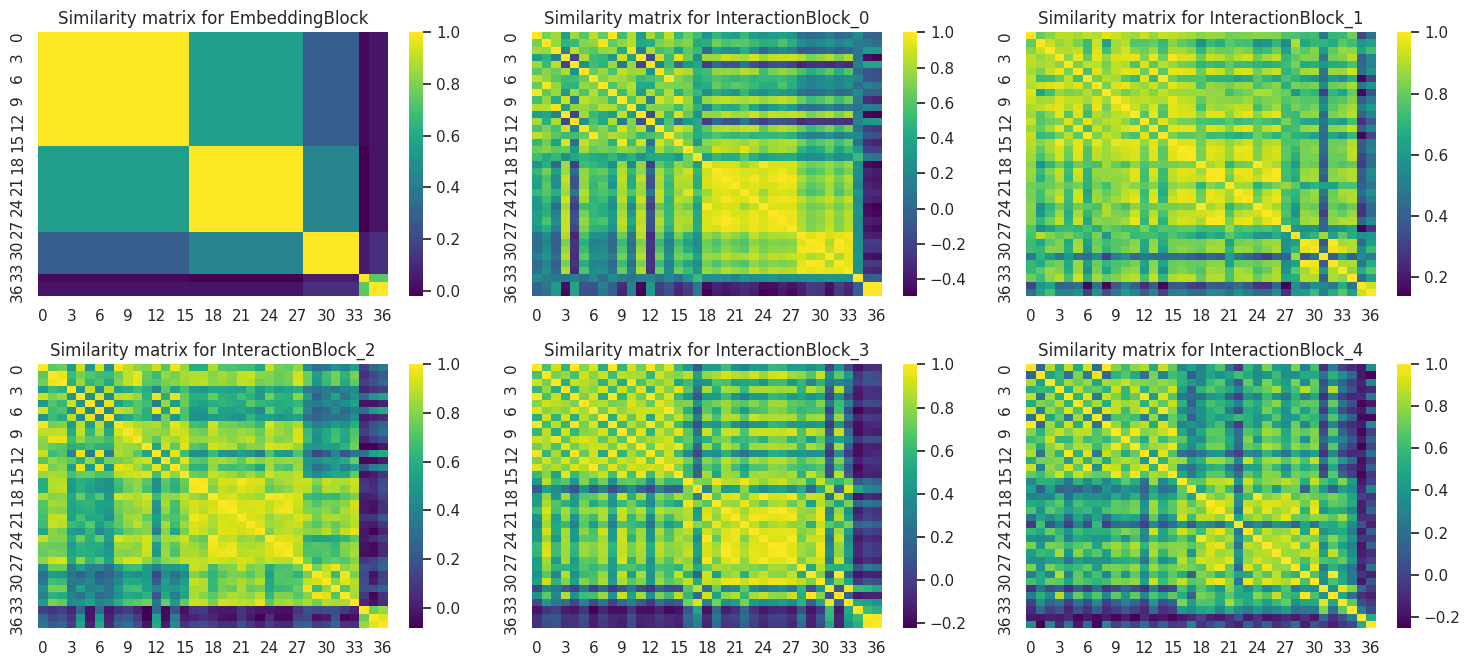}} \\
    \subfloat[SchNet with Ewald - Ewald Blocks]{\includegraphics[width=0.48\linewidth]{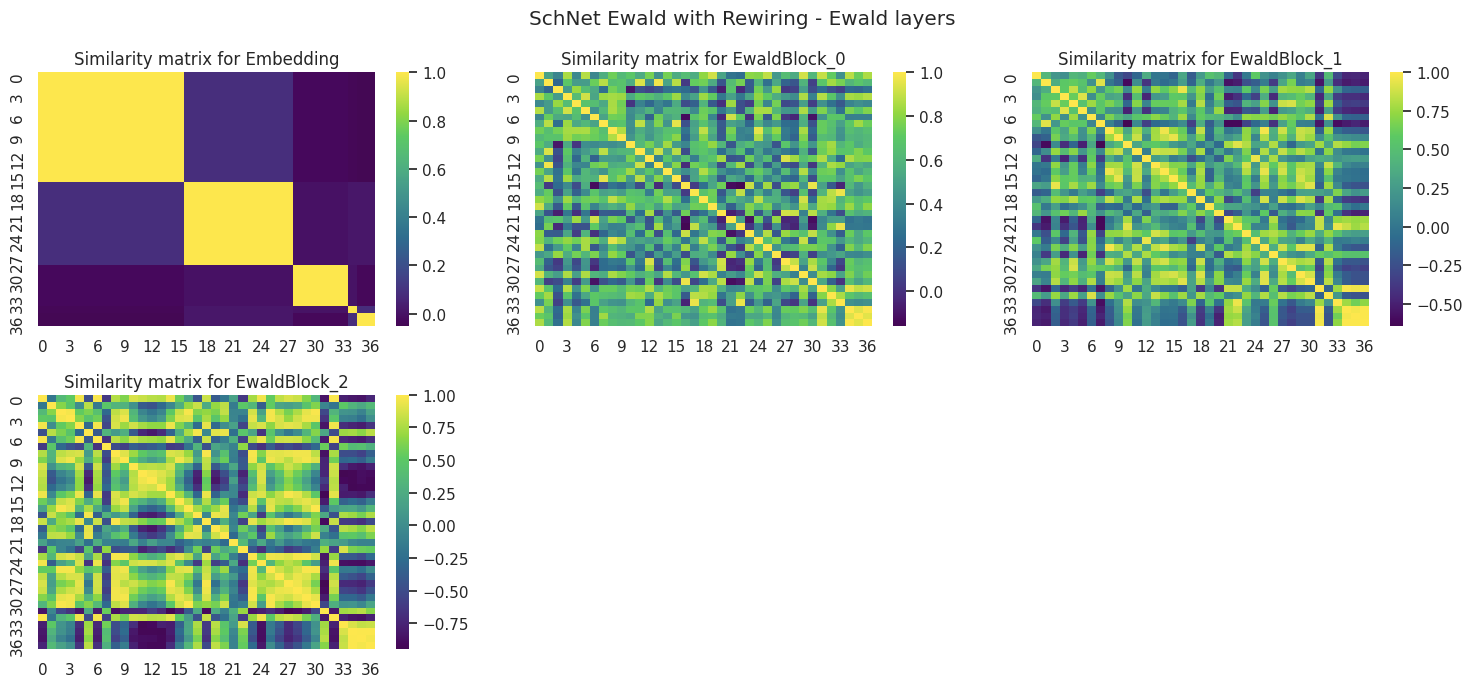}}
    \hfill
    \subfloat[FAENet with Ewald - Ewald Blocks]{\includegraphics[width=0.48\linewidth]{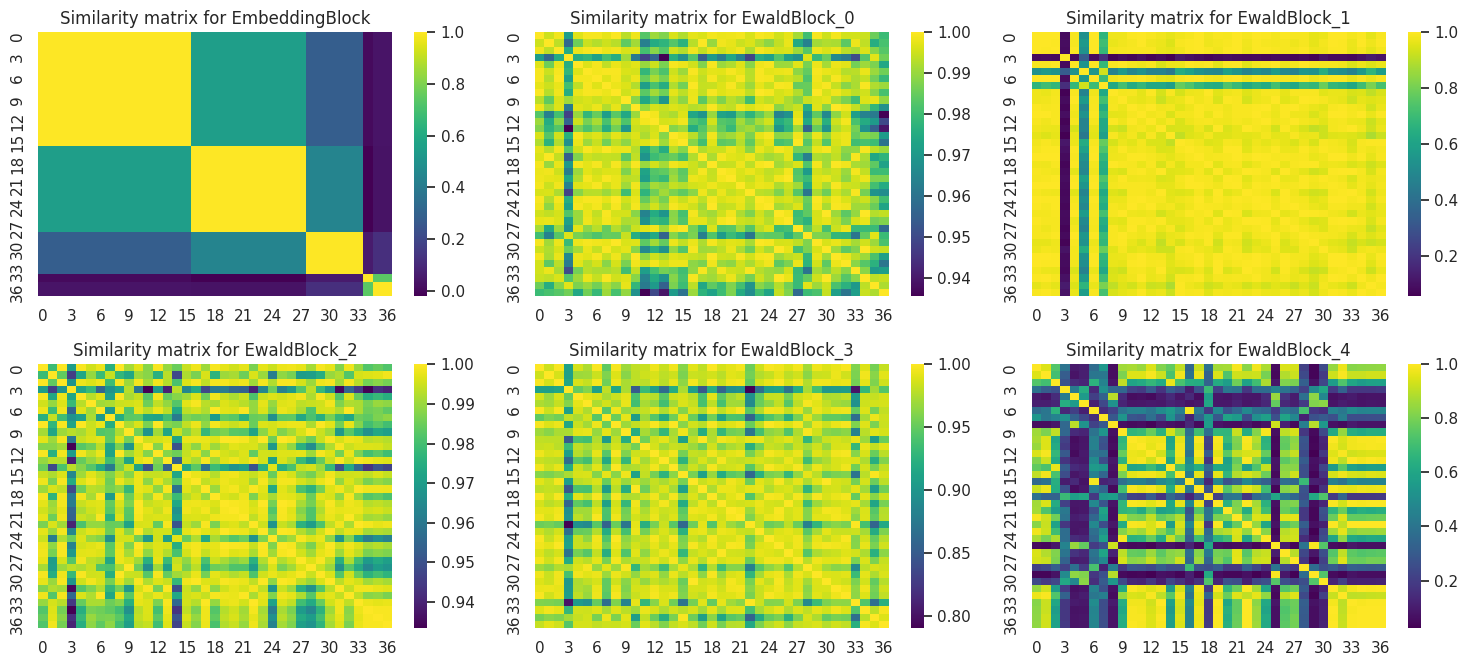}}
    \caption{Same plots as Figure~\ref{fig:ewald_embeddings_diverse} but with a second randomly picked system from the OC20 train split.}
    \label{fig:act_ewald_base2}
\end{figure}


\begin{table}[H]
\centering
\resizebox{1\textwidth}{!}{
    \begin{tabular}{l|cc|cc|cc|cc}
    & \multicolumn{2}{c|}{ID} & \multicolumn{2}{c|}{OOD-ADS} & \multicolumn{2}{c|}{OOD-CAT} & \multicolumn{2}{c}{OOD-BOTH} \\
    Model & EwT (\%) $\uparrow$& MAE (eV) $\downarrow$& EwT (\%) $\uparrow$& MAE (eV) $\downarrow$ & EwT (\%) $\uparrow$ & MAE (eV) $\downarrow$& EwT (\%) $\uparrow$& MAE (eV) $\downarrow$\\
    \hline
        FAENet (Graph Rewiring) & 4.05&  0.551 &  2.65&  0.650 &  4.29&  0.550&  2.76&  0.601 \\ 
        FAENet (Graph Rewiring) + Ewald & 4.12  & 0.562 &  2.68 &  0.648&  4.14&  0.563&  2.83& 0.597 \\ 
        FAENet (No Graph Rewiring) & 4.54 &  0.544 &  2.59&  0.657 &  4.66&  0.539&  2.65 & 0.601  \\ 
        FAENet (No Graph Rewiring) + Ewald & 4.11 & 0.556 & 2.75 & 0.626 & 4.13& 0.553 & 2.85 & 0.569 \\ 
        SchNet (Graph Rewiring) & 3.18 &  0.641 &  2.53&  0.720 &  3.00&  0.638&  2.59 & 0.642  \\ 
        SchNet (Graph Rewiring) + Ewald & 3.54 & 0.604 & 2.53 & 0.665 & 3.53& 0.599 & 2.67 & 0.608 \\ 
        SchNet (No Graph Rewiring) & 2.93&  0.654 &  2.22&  0.700 &  3.04&  0.646&  2.54&  0.656 \\ 
        SchNet (No Graph Rewiring) + Ewald & 3.48  & 0.597 &  2.76 &  0.647&  3.56&  0.599&  2.73& 0.612 \\ 
    \end{tabular}
    }
    \caption{Energy prediction errors with and without Ewald Message Passing. Graph Rewiring refers to removing the subsurface atoms from the system~\citep{duval2022phast}. In this table, FAENet is taken with 5 interaction layers (top config), while SchNet uses 3 interaction layers.}
\label{table:ewald_app}
\end{table}

\begin{table}[H]
\centering
\resizebox{0.45\textwidth}{!}{
    \begin{tabular}{l|cc}
    Model & \multicolumn{1}{c|}{MAE (meV) $\downarrow$} & \multicolumn{1}{c}{MSE ((meV)$^2$) $\downarrow$} \\
    \hline
        FAENet & 8.44 & 0.600 \\ 
        FAENet + Ewald & 8.34 & 0.574 \\ 
        SchNet & 16.0 & 1.18 \\ 
        SchNet + Ewald & 11.5 & 0.73 \\ 
    \end{tabular}
    }
    \caption{Energy prediction errors with and without Ewald Message Passing on the test split of the QM9 dataset for the target property $U_0$ (Internal energy at 0K).}
\label{table:ewald_app_qm9}
\end{table}

\section{Noisy Nodes}
\label{Noisy Nodes appendix}
\subsection{Experimental setup}
\label{Noisy Nodes appendix experimental setup}
Runs are done on a single Nvidia Quadro RTX 8000 GPU with 48 GB memory or, if indicated, on a single Nvidia A100 GPU with 80 GB memory.

\subsection{Noisy Nodes implementation}
\label{Noisy Nodes appendix implementation}
In practice, we perturb the input node positions of the graph $G$ with a noise $\sigma$ and train the model with two loss terms, a Noisy Nodes loss term and the primary loss (associated with the main task) term
\begin{equation}\label{Noisy nodes loss}
\mathcal{L} = \lambda \cdot \mathcal{L}_{NNodes} (\hat{G}',V')+\mathcal{L}_{Primary}(\hat{G}',V'),
\end{equation} 
where $\lambda$ is the weight we assign to the auxiliary denoising task, $\hat{G}'=\tilde \Phi(\tilde{G})$ is the output of the model $\tilde \Phi$, $\tilde{G}$ is the noised graph, and $V'$ can either be the target nodes features (e.g. atom positions at equilibrium, that is the IS2RS task) or the initial nodes positions (see next subsection on denoising pre-training). 

For the input, we first interpolate between initial
structure and relaxed structure and then add Gaussian noise, that is for each node $i$, the input positions of the input "Noisy Nodes graph" $\tilde{x}^i$ are defined as
\begin{equation}
\tilde{x}^i=
\begin{cases}
    \gamma(x^i_{rel}-x^i_{init})+Z^i  \textrm{  with probability 0.5} \\
    x^i_{init} \textrm{  with probability 0.5},
\end{cases}
\end{equation}
with random interpolation factor $\gamma\sim U[0,1]$ independent between graphs and iid Gaussian noises $Z^i\sim \mathcal{N}(0,\sigma)$ with $\sigma=0.3$.
The Noisy Nodes target is $\Delta_{pos}^i=x_{rel}^i-\tilde{x}^i$ and therefore the auxiliary loss term is $\|\Delta_{pos}^i-\Phi(\tilde{x}^i)\|_1$ for the model $\Phi$.
Our total loss is the sum of the energy MAE loss and the auxiliary loss weighted by a number $\lambda$, as in Equation~\ref{Noisy nodes loss}. Both the primary loss and Noisy Nodes Loss (before multiplication by  $\lambda$) typically have the same value between 1 and 2.5.

For the IS2RE training, we add a position decoding head to the preexisting energy prediction head. The position decoding head, as with the force prediction head introduced in \cite{duval2023faenet}, is a 2-layer MLP with Batchnorm.

\subsection{Related work}
Since Noisy Nodes performs denoising as an auxiliary task during training, the representation learning benefits of denoising are limited to the downstream dataset on which the model is trained. \citet{zaidi2022pretraining} propose to rather perform denoising as a pre-training objective on a large, unlabelled dataset of atomic structures.

\citet{shoghi2024JMP} introduce Joint
 Multi-domain Pre-training (JMP), a supervised pre-training strategy that simultaneously trains on various datasets from different chemical domains (OC20 \cite{chanussot2020open}, OC22 \cite{OC22}, ANI-1x \cite{ANI-1X},
 and Transition-1x \cite{schreiner2022transition1x}), treating
 each dataset as a unique pre-training task within a multi-task framework.

\subsection{Training hyperparameters}
\label{Noisy Nodes appendix training hyperparameters}

The most obvious changes to the training hyperparameters that should theoretically allow to leverage the denoising auxiliary task are to increase the depth of the network and the number of epochs. First, we observe that we reach convergence on the validation set for the energy MAE more slowly when using the auxiliary task because of a more complex loss to minimize and a higher number of model weights (due to the supplementary interaction blocks). Hence, unless otherwise stated, the number of epochs is 50 in the IS2RE with IS2RS auxiliary task experiments.

\cite{liao2023equiformer} use a linearly decaying weight from 15 to 1 for the auxiliary IS2RS loss to encourage the model to learn more from the auxiliary task in the beginning but focus on the primary task at the end of training. 
We also tested a constant auxiliary weight of 1, a weight decaying from 30, and cosine annealing with a linear warmup scheduler, but this yielded worse or equivalent results. Thus,
unless otherwise stated, we always use as an auxiliary weight scheduler the one of Equiformer in the following IS2RE with IS2RS auxiliary task experiments.

Since we always use the MAE as energy loss for the IS2RE main task, it proves essential to use the MAE loss for the auxiliary position loss to leverage the benefits of Noisy Nodes. Indeed, our experiments using the MSE position loss led to a collapse of the node embeddings at the last interaction layers, that could be observed by plotting the MAD throughout the interaction layers.

\subsection{Results}
\label{Noisy Nodes appendix results}

\subsubsection{Comparing Depth for classical FAENet}
In the results of Table~\ref{table:is2re_top_comp_interactions_fixed}, the number of warmup steps is 6000 for about 180k steps. The batch size is 128, hence the lower throughput than the one with the same configs of \cite{duval2023faenet}.
In these runs, the hyperparameters are the top configs of \cite{duval2023faenet} except for slight differences: the number of hidden channels in the embedding blocks is a bit lower. Also, contrary to the top configs of \cite{duval2023faenet}, we train for 50 epochs (instead of 12) with no early stopping to be in a comparable setup to the IS2RE with IS2RS auxiliary task models.
\begin{table*}[ht]
\centering
\resizebox{1\textwidth}{!}{
    \begin{tabular}{l|c|cc|ccccc|ccccc}
    Interaction & Parameters & \multicolumn{2}{c|}{Time} & \multicolumn{5}{c|}{Energy MAE (meV) $\downarrow$} & \multicolumn{5}{c}{EwT (\%) $\uparrow$}\\
     blocks & (millions) & Train $\downarrow$ & Infer. $\uparrow$ & ID & OOD Ads & OOD Cat & OOD Both & Average & ID & OOD Ads & OOD Cat & OOD Both & Average \\
    \hline
    5 & 5.9 & 19min & 786 & 556 & 685 & 552 & 636 & 607 & 4.41 & 2.26 & 4.51 & 2.47 & 3.41 \\ 
    8 & 9.2 & 25min & 676 & 554 & 643 & 558 & 596 & \textbf{588} & 4.31 & 2.66 & 4.40 & 2.75 & \underline{3.53} \\ 
    10 & 11.4 & 29min & 621 & 552 & 649 & 554 & 603 & \underline{590} & 4.56 & 2.77 & 4.20 & 2.72 & \textbf{3.56} \\ 
    12 & 13.7 & 34min & 527 & 551 & 661 & 556 & 609 & 594 & 4.01 & 2.57 & 4.21 & 2.53 & 3.33 \\ 
    14 & 15.9 & 37min & 498 & 621 & 749 & 608 & 693 & 668 & 3.42 & 2.06 & 3.27 & 2.31 & 2.77 \\ 
    16 & 18.2 & 40min & 457 & 590 & 704 & 592 & 647 & 633 & 4.05 & 2.51 & 3.90 & 2.53 & 3.25 \\ 
    \end{tabular}
    }
\caption{Comparison varying the number of interaction blocks for FAENet without auxiliary task with the top configs of \cite{duval2023faenet} except for slight differences. Scalability is measured with training time for one epoch (Train, in minutes) and inference throughput (Infer., number of samples processed in a second). The best score is in bold and the second-best score is underlined. Performances are slightly worse than the best performances presented in \cite{duval2023faenet} because of different hyperparameters, but the main conclusion is that the performances worsen when we add interaction layers after 12.}
\label{table:is2re_top_comp_interactions_fixed}
\end{table*}

\subsubsection{Comparing depth for FAENet with Noisy Nodes IS2RS auxiliary task}
In Table~\ref{table:is2re_aux_mae_comp_interactions_5_28}, we observe a clear positive correlation between the depth of the model and its performances in terms of energy MAE and EwT. The gain in performance is very clear between 5 and 16 interaction layers, then increases much more slowly up to 28 layers.


\begin{table*}[ht]
\centering
\resizebox{1\textwidth}{!}{
    \begin{tabular}{l|c|cc|ccccc|ccccc}
    Interaction & Parameters & \multicolumn{2}{c|}{Time} & \multicolumn{5}{c|}{Energy MAE (meV) $\downarrow$} & \multicolumn{5}{c}{EwT (\%) $\uparrow$}\\
    blocks & (millions) & Train $\downarrow$ & Infer.$\uparrow$ & ID & OOD Ads & OOD Cat & OOD Both & Average & ID & OOD Ads & OOD Cat & OOD Both & Average \\
    \hline
    5 & 4.2 & 17min & 724 & 523 & 592 & 525 & 545 & 546 & 4.59 & 2.92 & 4.40 & 3.15 & 3.76 \\ 
    8 & 6.6 & 22min & 679 & 518 & 600 & 526 & 561 & 551 & 5.13 & 2.88 & 5.21 & 2.74 & 3.99 \\ 
    10 & 8.1 & 27min & 584 & 513 & 606 & 521 & 561 & 550 & 5.17 & 2.82 & 5.03 & 2.81 & 3.96 \\ 
    12 (A100) & 9.7 & 30min & 504 & 517 & 589 & 523 & 546 & 544 & 5.10 & 3.17 & 4.92 & 3.04 & 4.06 \\ 
    14 & 11.2 & 33min & 525 & 507 & 580 & 519 & 541 & 537 & 5.20 & 3.13 & 5.21 & 3.25 & 4.20 \\ 
    16 & 12.8 & 35min & 465 & 505 & 566 & 516 & 527 & \underline{529} & 5.17 & 3.63 & 5.00 & 3.69 & 4.37 \\ 
    
    18 & 14.3 & 38min & 446 & 508 & 596 & 518 & 554 & 544 & 5.31 & 3.22 & 5.58 & 2.97 & 4.27 \\ 
    20 & 15.9 & 42min & 423 & 508 & 593 & 513 & 553 & 542 & 5.39 & 3.16 & 5.18 & 2.88 & 4.15 \\ 
    22 & 17.4 & 45min & 401 & 507 & 569 & 515 & 528 & 530 & 5.24 & 3.26 & 5.06 & 3.32 & 4.22 \\ 
    24 & 19.0 & 48min & 375 & 500 & 574 & 510 & 534 & \underline{529} & 5.70 & 3.43 & 5.43 & 3.31 & \textbf{4.47} \\ 
    26 & 20.6 & 50min & 335 & 504 & 562 & 512 & 521 & \textbf{525} & 5.15 & 3.68 & 5.20 & 3.67 & \underline{4.43} \\ 
    28 & 22.1 & 56min & 349 & 505 & 567 & 513 & 517 & \textbf{525} & 5.20 & 3.64 & 5.27 & 3.28 & 4.35 \\ 
    \end{tabular}
    }
\caption{Comparison of number of interaction blocks for IS2RE with auxiliary IS2RS. Scalability is measured witht raining time for one epoch (Train, in minutes) and inference throughput (Infer., number of samples processed in a second). Here the canonicalization technique is SE(3)-SFA, the number of warmup steps is 192000 (about half total number of steps), and the batch size is 64. Best score is in bold, second best score is underlined.}
\label{table:is2re_aux_mae_comp_interactions_5_28}
\end{table*}

\subsubsection{Canonicalization comparison}
We use SE(3)-SFA because it is less equivariant to reflections than Stochastic FA (which samples a frame in E(3)), but it has to learn data symmetries from fewer frames, which helps training.
For both SE(3)-SFA and No-FA models, we observe in Table~\ref{table:is2re_aux_compFA_test_ri} that the equivariance-invariance do not seem to be correlated to the number of interaction layers. Moreover, we do not observe a clear correlation between the Average energy MAE and the equivariance-invariance. 

When using SE(3)-SFA, we see a clear correlation between the increase in the number of interaction layers and the performances. In the No-FA case, the impossibility to learn equivariance and invariance even when adding more layers to the model might account for the less clear positive correlation between the performances (in terms of energy MAE) and the number of interaction layers than when we were applying SE(3)-SFA.
\begin{table*}[htbp]
\centering
\resizebox{0.9\textwidth}{!}{
    \begin{tabular}{l|c|cccc|c}
    Canonicalization & Interactions & 2D E-RI $\downarrow$ & 2D E-Refl-I $\downarrow$ & 2D Pos-RI $\downarrow$ & 2D Pos-Refl-I $\downarrow$ & Average E-MAE (meV) $\downarrow$ \\
    \hline
    SE(3)-SFA & 5 & 22.7 & 34.7 & 55.0 & 80.3 & 546 \\ 
    SE(3)-SFA & 14 & 29.7 & 43.2 & 61.7 & 90.4 & 537 \\ 
    SE(3)-SFA & 16 & 21.4 & 33.6 & 56.0 & 83.4 & \underline{529} \\ 
    SE(3)-SFA & 18 & 19.0 & 30.2 & 53.9 & 78.4 & 544 \\ 
    SE(3)-SFA & 20 & 22.6 & 34.7 & 56.2 & 82.8 & 542 \\ 
    SE(3)-SFA & 22 & 20.8 & 32.6 & 56.3 & 81.7 & 530 \\ 
    SE(3)-SFA & 28 & 25.5 & 37.3 & 60.8 & 86.5 & \textbf{525} \\ 
\hline
    SE(3)-SFA no aux & 5 & 6.9 & 8.9 &  &  & 569 \\ 
\hline
    No-FA & 8 & 111  & 107 & 253 & 238 & \underline{561} \\ 
    No-FA & 10 & 110  & 107 & 251 & 233 & 577 \\ 
    No-FA & 14 & 121  & 117 & 275 & 254 & 562 \\ 
    No-FA & 18 & 115  & 111 & 273 & 257 & 608 \\ 
    No-FA & 22 & 116  & 111 & 260 & 240 & \textbf{554} \\ 
    No-FA & 26 & 120  & 117 & 288 & 271 & 579 \\ 
    \end{tabular}
}
\caption{Comparison of using SE(3)-SFA on both the energy and position prediction heads to No FA and to the top configs of FAENet with no auxiliary task from \cite{duval2023faenet}. The symmetry-preservation metrics are in meV for the energy and milli-Angstroms for the positions. The best model for each of the 2 categories is in bold and the second best is underlined. We do not seem to have a correlation between the quality of the invariance and equivariance with the number of interaction layers in both categories.}
\label{table:is2re_aux_compFA_test_ri}
\end{table*}

\subsection{Pre-tranining on different tasks}
\label{sec:app_ft_is2re}
    \begin{figure}[ht]
        \centering
        \includegraphics[width=0.7\linewidth]{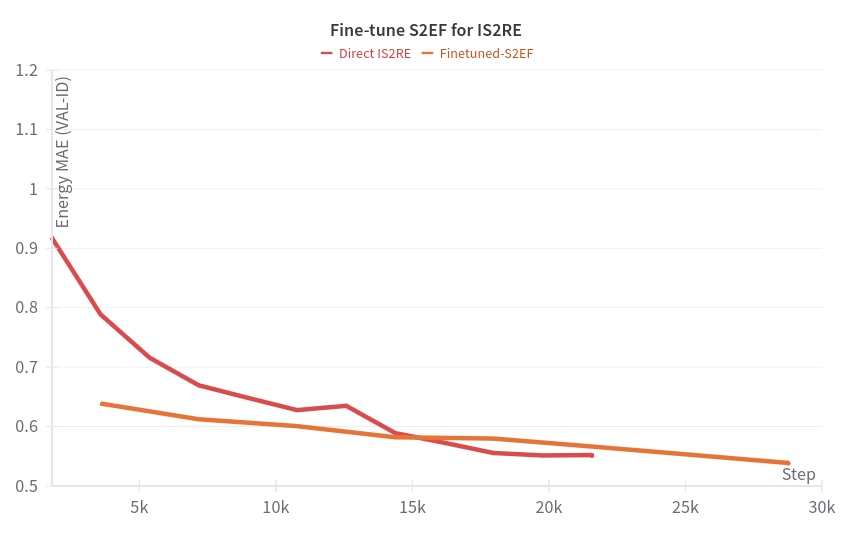}
        \caption{Validation curve during training for both a model trained directly from scratch for IS2RE and an S2EF model fine-tuned on IS2RE}
        \label{fig:ft-is2re}
    \end{figure}

%



\end{document}


\twocolumn[
\icmltitle{Improving molecular modeling with geometric GNNs: an empirical review}



\icmlsetsymbol{equal}{*}

\begin{icmlauthorlist}
\icmlauthor{Aeiau Zzzz}{equal,to}
\icmlauthor{Bauiu C.~Yyyy}{equal,to,goo}
\end{icmlauthorlist}

\icmlaffiliation{to}{Department of Computation, University of Torontoland, Torontoland, Canada}
\icmlaffiliation{goo}{Googol ShallowMind, New London, Michigan, USA}
\icmlaffiliation{ed}{School of Computation, University of Edenborrow, Edenborrow, United Kingdom}

\icmlcorrespondingauthor{Cieua Vvvvv}{c.vvvvv@googol.com}
\icmlcorrespondingauthor{Eee Pppp}{ep@eden.co.uk}

\icmlkeywords{Machine Learning, ICML}

\vskip 0.3in
]



\printAffiliationsAndNotice{\icmlEqualContribution} 

\section{Noisy Nodes Experimental setup}
\label{Noisy Nodes Experimental setup}

\nocite{langley00}

\bibliography{main}
\bibliographystyle{icml2021}